\documentclass[10pt,twocolumn,letterpaper]{article}

\usepackage[pagenumbers]{cvpr} %
\usepackage{chen}

\def\TASK{Visual Abductive Reasoning}
\def\TASKAc{VAR}
\def\OURS{{\textsc{Reasoner}}}
\def\DATASET{VAR}

\usepackage[pagebackref,breaklinks,colorlinks]{hyperref}

\def\cvprPaperID{}
\def\confName{CVPR}
\def\confYear{2022}

\definecolor{myred}{RGB}{190,0,0}
\definecolor{myblue}{RGB}{30,80,120}

\begin{document}

\title{Visual Abductive Reasoning}

\author{Chen Liang$^{1,4}$\thanks{Part of this work was done when Chen Liang was an intern at Baidu.}~~, Wenguan Wang$^{2}$~, Tianfei Zhou$^{3}$~, Yi Yang$^1$\\
\small{$^1$CCAI, Zhejiang University~~~$^2$ReLER, AAII, University of Technology Sydney~~~$^3$ETH Zurich~~~$^4$Baidu Research}\\
\small\url{https://github.com/leonnnop/VAR}
}

\maketitle

\begin{abstract}

Abductive reasoning seeks the likeliest possible explanation for partial observations. Although abduction is frequently employed in human daily reasoning, it is rarely explored in computer vision literature. In this paper,~we~propose a new task and dataset, Visual Abductive Reasoning (VAR), for examining abductive reasoning ability of ma- chine$_{\!}$   intelligence$_{\!}$  in$_{\!}$  everyday$_{\!}$  visual$_{\!}$  situations.$_{\!}$  Given$_{\!}$ an$_{\!}$ in- complete set of visual events, AI systems are required to not\\
 only describe what is observed, but also {infer} the hypothe- sis that can best explain the visual premise.$_{\!}$ Based on our large-scale VAR dataset, we devise a strong baseline model,$_{\!}$ {\OURS}$_{\!}$ (causal-and-cascaded reasoning Transformer). First, to capture the {causal structure} of the observations, a\\
 contextualized$_{\!}$ directional$_{\!}$ position embedding strategy is adopted in the encoder, that yields discriminative representations for the premise and hypothesis.$_{\!}$ {Then, multiple decoders are cascaded to generate and progressively refine the premise and hypothesis sentences.} The prediction scores of the sentences are used to guide cross-sentence information flow in the cascaded reasoning {procedure}. Our VAR benchmarking results show that {\OURS} surpasses many famous video-language models, while still being far behind human performance.
This work is expected to foster future efforts in the reasoning-beyond-observation paradigm.

\end{abstract}

\vspace{-12pt}
\section{Introduction}
\vspace{-3pt}

\begin{quote}
\it \small
Abduction $\cdots$ consists in studying facts and devising a theory
to explain them. \\
\mbox{}\hfill -- Charles Sanders Peirce (1839 -- 1914)
\vspace{-3pt}
\end{quote}

\begin{figure}[t]
    \vspace{-1.5ex}
    \begin{center}
        \includegraphics[width=0.95\linewidth]{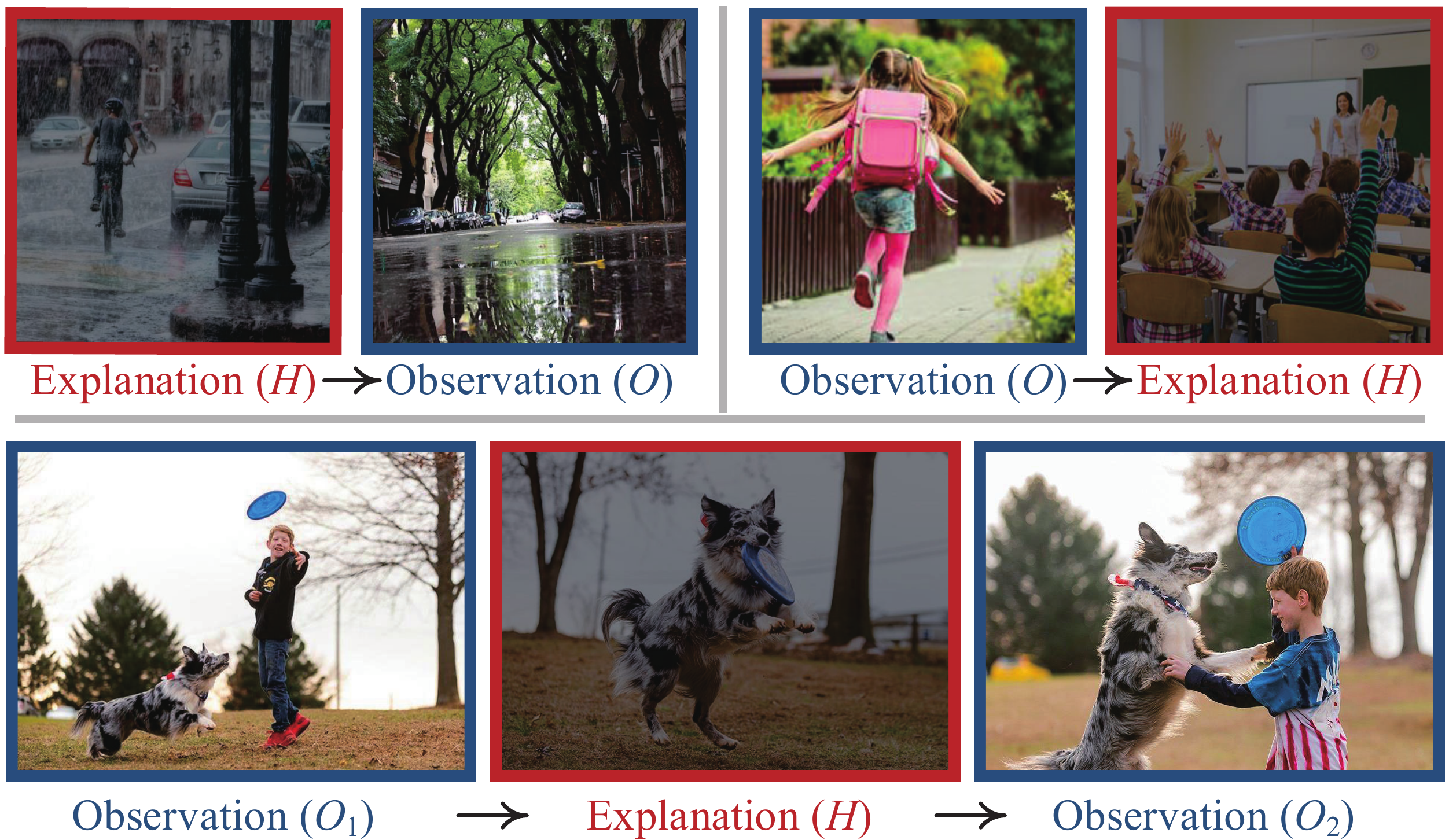}
    \end{center}
    \vspace{-18pt}
	\captionsetup{font=small}
	\caption{\small{Abductive reasoning is inference to the most likely {\color{myred}{explanation}} for an incomplete set of {\color{myblue}{observations}}.}}
	\vspace{-17pt}
	\label{fig:motivation}
\end{figure}

\textit{Abductive reasoning}~\cite{peirce1931collected} was coined by Charles Sanders Peirce, the founder of American pragmatism, around 1865. It is inference to the most likely explanation or conclusion for an incomplete set of observations. Abductive reasoning is invariably employed in our everyday life; the generated hypothesis (${\color{myred}{H}}$) is expected to best explain what happens before, after, or during the observation (${\color{myblue}{O}}$). Fig.~\ref{fig:motivation} gives some examples. If you~see ${\color{myblue}{O}}$: ``{\color{myblue}{the road is wet}}'', abduction will lead you to the best explanation ${\color{myred}{H}}$: ``{\color{myred}{it rained earlier}}'' (\ie, ${\color{myred}{H}}_{\!}\!\rightarrow_{\!}\!{\color{myblue}{O}}$). One morning you find ${\color{myblue}{O}}$: ``{\color{myblue}{sister leaves home hurriedly}}'', then you conclude ${\color{myred}{H}}$: ``{\color{myred}{she will be late for class}}'' (\ie, ${\color{myblue}{O}}\!\rightarrow\!{\color{myred}{H}}$).
You see ${\color{myblue}{O_1}}$: ``{\color{myblue}{a boy throws a frisbee out and his dog is running after it}}''. One minute later you find ${\color{myblue}{O_2}}$: ``{\color{myblue}{frisbee is in the boy's hand}}''. You~can imagine ${\color{myred}{H}}$:$_{\!}$ ``{\color{myred}{the$_{\!}$ dog$_{\!}$ just$_{\!}$ caught$_{\!}$ the$_{\!}$ frisbee$_{\!}$ back}}''$_{\!}$ (\ie, ${\color{myblue}{O_{1\!}}}\!\rightarrow_{\!}\!{\color{myred}{H}}_{\!}\!\rightarrow\!{\color{myblue}{O_2}}$). Although abductive reasoning has long been considered as a core ability of everyday human cognition~\cite{lombrozoexplanation, shelley1995visual, shanahan2005perception}, it still remains an untouched domain in computer vision literature.

In this article, we propose \textit{Visual Abductive Reasoning} ({\TASKAc}), a novel task and dataset for investigating the abductive reasoning ability of AI systems in daily visual situations. In particular, inspired by the recent advance of causal reasoning in NLP community (\ie, abductive text generation~\cite{bhagavatula2020abductive} and counterfactual story revision \cite{qin2019counterfactual}), we explore the use of natural language as the expression form to fully capture the complexity of real situations. This also better reflects the nature of human mind, which relies on linguistic thinking$_{\!}$~\cite{logan2018topology,logan2018thinking}. {\TASKAc} requires  the AI systems to
describe the incomplete observation (\ie, visual premise) and write down the hypothesis that can best explain the premise.  This allows to thoroughly evaluate the entire abduction procedure, as accurate understanding of the premise is the basis of abductive reasoning. Moreover, this can hasten the development of this new field, by comparing and
embracing ideas for a relevant, well-established, yet different task
-- dense video captioning (DVC)$_{\!}$~\cite{krishna2017dense}. In contrast to DVC that focuses only on \textit{describing the observation}, {\TASKAc}{\ms}yields{\ms}a{\ms}new{\ms}visual-linguistic{\ms}reasoning{\ms}paradigm -- \textit{inference beyond observation}. Three characteristics make {\TASKAc} uniquely challenging: \textbf{i)} {\TASKAc}{\ms}needs{\ms}imagination{\ms}to{\ms}find{\ms}hypotheses{\ms}\textit{outside} the observation. \textbf{ii)} {\TASKAc} seeks to discover the plausible causal structure among the observed events. \textbf{iii)} {\TASKAc} is more related to the kind of human reasoning in daily situations where the information at hand is {often incomplete~\cite{keil2003folkscience}} and absolutely certain conclusions cannot be reached$_{\!}$~{\cite{bhagavatula2020abductive,keil2006explanation}.}

Our dataset is collected to address the characteristics of the {\TASKAc} task (\textit{cf}.$_{\!}$~\S\ref{sec:td}). It
contains \cnum{9}K examples from \cnum{3718} videos. Each example consists of several chronologically-ordered events, most of which are logically related. For each event, abduction oriented description is written by people, and its role of \textit{premise} or \textit{explanation} is also annotated. When presenting each example to the AI system, the {explanation} event is masked and premise events are visible. The AI system is required to understand the partial, noisy observations (\ie, premise events) and construct the most plausible explanatory hypothesis -- accurately describing both the observable premise events and the masked explanation event. The examples are on average \cnum{37.8}s long, with \cnum{4.17} events, and harvested from diversely event-rich sources, \ie, YouTube Lifestyle videos, movies and TV shows.

To lay a solid foundation for future efforts, a new model, named {\OURS} (causal-and-cascaded reasoning Transformer), is proposed (\textit{cf}.$_{\!}$~\S\ref{sec:md}).  Specifically, {\OURS} is building upon a Transformer encoder-decoder architecture. In the encoder of {\OURS}, a \textit{contextualized} \textit{directional$_{\!}$ position$_{\!}$ embedding}$_{\!}$ strategy$_{\!}$ is adopted to$_{\!}$ capture$_{\!}$ causal dependencies among the premise events. Hence the context of the premise events can be gathered in a causality-aware manner, enabling {\OURS} to learn discriminative representations for the premise and explanatory hypothesis. Then {\OURS} cascades a set of decoders for premise/hypothesis sentence production and refinement. For each generated sentence, the associated prediction score is viewed as the confidence and embedded into the next decoder as a signal for inspiring more information to flow from high-confident sentences to the low-confident ones. This leads to a \textit{confidence-guided multi-step reasoning} strategy, boosting the reasoning power of {\OURS} eventually.

Extensive experimental results are provided in \S\ref{sec:exp}. First, to comprehensively probe deep neural models on this challenging task, we establish a group of baselines based on current top-leading DVC models. The benchmarking results on {\TASKAc} dataset show that {\OURS} outperforms the best competitor by a large margin, \eg, {\cnum{33.44}} \textit{vs} {\cnum{28.71}} in terms of BERT-S, but is still far behind human performance ({\cnum{42.96}}). This shows that VAR is especially challenging for current video-language models.  Then a set of user studies and ablative experiments are conducted for a thorough evaluation. For completeness, we further test {\OURS} on the DVC task and confirm again its superiority.

$_{\!}$Concurrent$_{\!}$ to$_{\!}$ us,$_{\!}$ \cite{hesselhwang2022abduction}$_{\!}$ studies$_{\!}$ \textit{image}-based$_{\!}$ abductive$_{\!}$~rea- soning: AI systems are required~to~identify, ground, or compare \textit{given} inferences. Overall, we feel vision-based abductive reasoning is an intriguing problem worthy of exploring.

\vspace{-3pt}
\section{Related Work}
\vspace{-1pt}
\paragraph{Dense Video Captioning (DVC).} Different from the classic video description task$_{\!}$~\cite{venugopalan2015translating,venugopalan2015sequence,yao2015describing,pan2016hierarchical,kojima2002natural,zhang2020object} that aims to describe a short video clip using a single sentence, DVC is to comprehensively describe all the events in an untrimmed video$_{\!}$ through$_{\!}$ a$_{\!}$ multi-sentence$_{\!}$ paragraph$_{\!}$~\cite{krishna2017dense}.$_{\!}$ Typical$_{\!}$ DVC
models$_{\!}$~\cite{krishna2017dense,zhou2018end,melas2018training,xiong2018move,zhou2019grounded,mun2019streamlined,pan2020spatial,song2021towards} follow a \textit{two-stage}, \textit{bottom-up} paradigm: first parse a video into several temporal events and then decode a description from each detected event. As the problem of event detection is ill-defined$_{\!}$~\cite{deng2021sketch}, some alterative solutions either adopt a \textit{single-stage} strategy to simultaneously predict events and descriptions \cite{li2018jointly,wang2021end}, or turn to a \textit{top-down} regime: first generate paragraphs, and then ground each description to$_{\!}$ a$_{\!}$ video$_{\!}$ segment$_{\!}$~\cite{deng2021sketch,liu2021video}.$_{\!}$ A$_{\!}$ few
other methods$_{\!}$~\cite{park2019adversarial,lei2020mart,ji2021hierarchical} focus purely on generating better paragraph captions from a provided list of events.

Both {\TASKAc} and DVC are concerned with video-based text generation; a part of our dataset is sourced from ActivityNet Captions~\cite{krishna2017dense}, a famous DVC dataset. However, DVC is aware of general fact based plain narrative, while {\TASKAc} addresses  cause-effect chain based abductive reasoning. Rather than accurately understanding what is observed, {\TASKAc} further requires {invoking} what might have happened or will happen. In our experiments, we involve several recent DVC models as baselines for our {\TASKAc} task and also report the performance of our {\OURS} on the DVC task.

\paragraph{Context-Aware Text Generation.} Our work is also related to some context-aware text generation tasks in the NLP literature. For instance, \textit{text infilling}$_{\!}$~\cite{zhu2019text}, also known as the \textit{cloze task}$_{\!}$~\cite{taylor1953cloze}, is to generate a span of missing tokens in a text chunk, while \textit{sentence/story} \textit{infilling}$_{\!}$~\cite{ippolito2019unsupervised,huang2020inset} aims to generate missing sentences in long-form text. The generated tokens/sentences are expected to smoothly blend into and fit the context syntactically$_{\!}$~\cite{zhu2019text}, semantically$_{\!}$~\cite{ippolito2019unsupervised,huang2020inset}, and$_{\!}$ logically$_{\!}$~\cite{kang2019linguistic}.$_{\!}$ \textit{Counterfactual$_{\!}$ story$_{\!}$ revision}$_{\!}$~\cite{qin2019counterfactual} requires generating$_{\!}$ a$_{\!}$ new$_{\!}$ ending,$_{\!}$ given$_{\!}$ a$_{\!}$ story$_{\!}$ context$_{\!}$ altered$_{\!}$
by a counterfactual condition. Our work draws inspiration from$_{\!}$ \textit{abductive text
generation}$_{\!}$~\cite{bhagavatula2020abductive}, which investigates abductive reasoning  via a natural language inference task: write an appropriate reason that could explain observations described by narrative text. Unlike these language tasks addressing inter-sentential relationship understanding only, our {\TASKAc} task requires abduction and narrative for a sequence of partially observable visual events. Moreover, our {\TASKAc} task setting is more general; it is not limited to the strict form of abductive reasoning in \cite{bhagavatula2020abductive}, \ie, generate a hypothesis ($H$) of what happened between the observed past ($O_1$) and future ($O_2$) contexts:  $O_1\!\rightarrow\!H\!\rightarrow\!O_2$, but involves $O\!\rightarrow\!H$ and $H\!\rightarrow\!O$ abductive reasoning cases.

\paragraph{Visual Future/State Prediction.} Our work is, to some degree, relevant to future prediction -- a popular research area in computer vision. In this area,
a huge spectrum of topics/tasks are put forward, including forecasting future frames \cite{mathieu2016deep,vondrick2016generating}, future features \cite{vondrick2016anticipating,suris2021learning},  future actions$_{\!}$~\cite{ryoo2011human,kitani2012activity,lan2014hierarchical,abu2018will,sun2019relational}, future human motions$_{\!}$~\cite{fragkiadaki2015recurrent,jain2016structural,martinez2017human}, future human trajectories$_{\!}$~\cite{alahi2016social}, future goals \cite{epstein2021learning}, \etc. Rather than studying the future generation at the semantic-category or color-pixel level, event-level prediction was recently addressed in$_{\!}$~\cite{lei2020vlep} and$_{\!}$~\cite{park2020visualcomet}. However, \cite{lei2020vlep} only requires choosing from two candidates for future event prediction, making the take less challenging. \cite{park2020visualcomet} targets to describe past, present, and future events for a single image, while our {\TASKAc} task requires making full use of the information from a set of premise events. There are also some efforts made towards understanding the dynamics/transformations between states$_{\!}$~\cite{park2019robust,hong2021transformation,yang2021multiple} or goal-conditioned procedure planning$_{\!}$~\cite{chang2020procedure}, while either relying on a pre-defined, limited action prediction space~\cite{hong2021transformation,chang2020procedure}, or using simulated environments~\cite{chang2020procedure,hong2021transformation}. Our {\TASKAc} task is essentially to discover and describe the causal relations in real visual environments. It is not constrained to a narrow view of predicting either future events or between-state changes, but tries to infer the missing parts in the cause-effect chains, even with some unrelated premise events. All of these together make {\TASKAc} a unique and challenging visual reasoning task.

\paragraph{Position Encoding in Transformers.} Due to the permutation invariant nature of the attention operation, \cite{shaw2018self} learns and encodes position embeddings into Transformer tokens. Subsequent language-Transformers hence explore further variations, like incorporating sinusoid prior with more parameters~\cite{dai2019transformer}, simplifying position embeddings as learnable scalars\tcite{raffel2020exploring}, disentangling special tokens$_{\!}$ ($_{\!}$\texttt{[CLS]}$_{\!}$)\tcite{ke2020rethinking}, \etc. Some recent vision-Transformers~\cite{wu2021rethinking,chu2021twins} consider directional relative distance between 2D positions, and/or the interactions between visual tokens and position embeddings. {\OURS} encodes the relations of premise events in a directional and contextualized manner for causal relation modeling, and leverages the prediction scores of descriptions for confidence-guided multi-step reasoning.

\vspace{-3pt}
\section{Our {\TASKAc} Task and Dataset}\label{sec:td}
\vspace{-2pt}
\subsection{{\TASKAc} Task}\label{sec:task_definition}
\vspace{-2pt}
Our visual abductive reasoning ({\TASKAc}) task is designed to test the abductive reasoning ability of machine intelligence in everyday visual environments $\mathcal{E}$. Formally, given a video example $\mathcal{V}\!\subset\!\mathcal{E}$ that contains  a set of  $N$ events, \ie, $\mathcal{V}\!=\!\{O_1, \cdots_{\!}, O_{n-1}, H, O_n, \cdots_{\!}, O_{N-1}\}$, which are logically related and chronologically organized, we denote $\{O_n\}_{n=1}^{N-1\!}$ as \textit{premise} events -- partial observation of $\mathcal{E}$, and $H$ as \textit{explanation} event that can best explain the premise events. Conditioning on the past \textit{and/or} future visual context $\{O_n\}_{n=1}^{N-1\!}$ \textit{only}, the AI system is asked to describe these premise events, and, more importantly, infer and write down the most plausible explanatory hypothesis for the premise. Naturally, such a hypothesis is expected to be consistent with the content of the {explanation} event $H$. The abduction ability can thus be thoroughly examined by assessing  the quality of both the premise-based descriptions and explanatory hypothesis sentences -- {adequent} understanding of the premise is a necessary prerequisite for abductive reasoning.

\begin{figure}[t]
    \begin{center}
        \includegraphics[width=1.\linewidth]{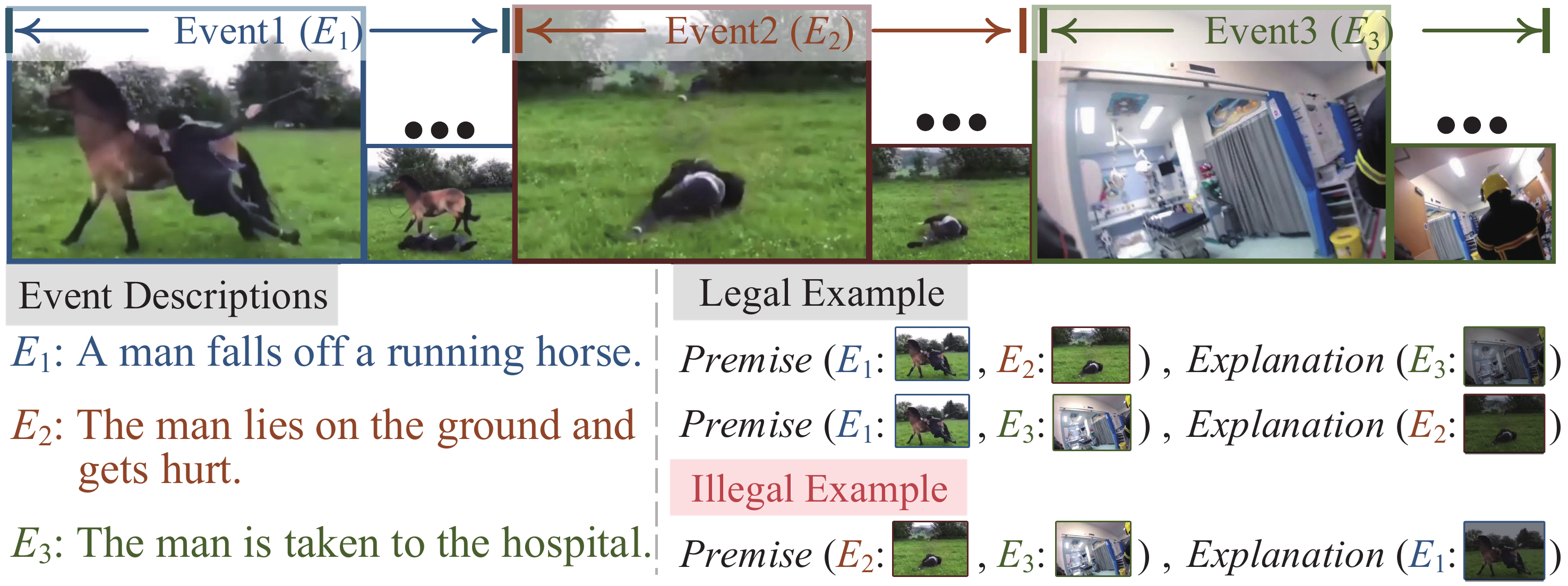}
    \end{center}
    \vspace{-18pt}
	\captionsetup{font=small}
	\caption{\small An illustrative example of our {\TASKAc} dataset (\S\ref{sec:dataset}).}
	\vspace{-14pt}
	\label{fig:dataset:example}
\end{figure}

\subsection{{\TASKAc} Dataset}\label{sec:dataset}
Guided by the above task setup, we build a large-scale dataset for {\TASKAc}. Fig.~\ref{fig:dataset:example} depicts an illustrative example.

\vspace{-10pt}
\subsubsection{Dataset Collection}
\vspace{-4pt}

\paragraph{Data Source.} {\DATASET} dataset is collected from three sources:
\begin{itemize}[leftmargin=*]
	\setlength{\itemsep}{0pt}
	\setlength{\parsep}{-2pt}
	\setlength{\parskip}{-0pt}
	\setlength{\leftmargin}{-10pt}
	\vspace{-6pt}
    \item \cnum{23457} \textit{Youtube} lifestyle Vlog videos from ActivityNet Captions~\cite{krishna2017dense} and VLEP~\cite{lei2020vlep} datasets. These videos cover rich social scenarios and human activities.
    \item {
            \cnum{13799} \textit{TV show and movie} videos from TVC dataset~\cite{lei2020tvr} and a famous Youtube channel, Fandango MovieClips\footnote{\url{https://youtube.com/user/movieclips}}.
            These videos are key scenes in popular TV shows and films containing wide-ranging genres.
        }
	\vspace{-6pt}
\end{itemize}
YouTube videos include  diverse daily events, but last relatively short durations with short intervals (about minutes). While TV shows and movie videos usually have limited scenarios,$_{\!}$ they$_{\!}$ contain$_{\!}$ rich$_{\!}$ artificial$_{\!}$ cause-effect chains in their story-lines and last relatively long durations with long intervals (spanning even years). Thus gathering these videos together makes our dataset a good testbed for {\TASKAc}.

\paragraph{Data$_{\!}$ Cleaning.$_{\!\!}$} The$_{\!\!}$ collected$_{\!\!}$ videos$_{\!\!}$ are$_{\!\!}$ accompanied$_{\!\!}$~by event labels, and{\ms}videos{\ms}containing{\ms}only{\ms}one{\ms}single event are first dropped. Then, for each of the rest videos, three human experts are invited to examine if there exists cause-effect relations between the video events. We only preserve qualified ones with more than two votes in the affirmative, finally resulting in \cnum{3718} videos in total for further annotation.

\vspace{-10pt}
\subsubsection{Dataset Annotation}
\vbox{ %
For each video in {\DATASET}, the annotation contains three steps:
\noindent\textbf{Step 1: Event Type Annotation.} For an event $E$ of video
}
\noindent $\mathcal{V}$, if $E$ can well explain some other events in $\mathcal{V}$; or in other words, if we can imagine that $E$ could happen by only considering the other events $\mathcal{V}/E$, event ${E}$ will be labeled as \textit{explanation} and the other events $\mathcal{V}/E$ will be labeled as \textit{premise}. 	Fig.~\ref{fig:dataset:example} gives an example. For the video containing three events: $E_1$ ``a man falls off a running horse'', $E_2$ ``the man lies on the ground and gets hurt'', $E_3$ ``the man is taken to the hospital'', we can derive two legal examples for our {\DATASET} dataset: \{\textit{premise} ($E_1$, $E_2$), \textit{explanation} ($E_3$)\}, and \{\textit{premise} ($E_1$, $E_3$), \textit{explanation} ($E_2$)\}. %

\paragraph{Step 2: Abductive Reasoning Aware Description Anno-  tation.$_{\!}$} Although$_{\!}$ some$_{\!}$ videos$_{\!}$ are$_{\!}$ collected$_{\!}$ with event-level descriptions/plot summaries, we re-annotate all the events with abductive reasoning$_{\!}$ oriented$_{\!}$ descriptions.$_{\!}$ Specifically, instead of capturing all the visual details in video captioning, like \{``a boy~in~a black jacket plays frisbee with his white dog in a park'', ``the dog catches the blue frisbee and runs back'', ``the boy smiles, takes the frisbee and pats the dog\}, our descriptions are only aware of describing the visual content related to abductive reasoning, like \{``a boy throws a frisbee out and his dog is running after it'', ``the dog catches the frisbee back'', ``the boy gets the frisbee''\}. %

\paragraph{Step 3: Validation.} Finally, each annotated example is examined by three verifiers: the verifiers are shown with only the \textit{premise} events and language-based explanation (\ie, description on the \textit{explanation} event), and vote for: ``Is the explanation sound?''. If an example wins majority approval, it will be accepted; otherwise, it will be relabeled or dropped.

\vspace{-12pt}
\subsubsection{Dataset Features and Statistics}\label{sec:dfs}
\vspace{-4pt}
To offer deeper insights into our {\DATASET} dataset, we next discuss its distinctive properties and detailed statistics.
\paragraph{Abductive Reasoning Orientated.}
{\DATASET} is the first dataset that underpins machine intelligence study of abductive reasoning in visual daily scenarios.
It is designed to reason beyond visual \textit{premise} for a plausible \textit{explanation}, distinguishing it from existing video-language datasets/tasks.

\paragraph{Diversity.} To capture diverse cause-effect relations and abduction cases,  our {\DATASET} dataset covers
\textbf{i)} various daily events/activities, \eg, work, leisure, household;
\textbf{ii)} rich scenarios, \eg, lifestyle recording, scripted drama;
\textbf{iii)} different durations and intervals, ranging from minutes to years.

\begin{figure}[t]
    \vspace{-4pt}
    \begin{minipage}{\textwidth}
        \hspace{-1ex}
        \begin{minipage}[t]{0.2\textwidth}
            \vspace{-11ex}
            \begin{threeparttable}[t]
                \small
                \resizebox{\linewidth}{!}{
                \setlength\tabcolsep{2pt}
                \renewcommand\arraystretch{1.1}
                \begin{tabular}{c|ccc}
                    \hline\thickhline
                    \rowcolor{mygray}
                    {Split} & {\#Example} & {\#Event} & {\#Video} \\
                    \hline \hline
                    \texttt{Train} & \cnum{7053} & \cnum{12582} & \cnum{3000} \\
                    \texttt{Val}   & \cnum{460} & \cnum{860}  & \cnum{205}\\
                    \texttt{Test}  & \cnum{1093} & \cnum{2044}  & \cnum{513}\\
                    \hline
                    \texttt{Total} & \cnum{8606} & \cnum{15486} & \cnum{3718} \\
                    \hline
                \end{tabular}
                }
            \end{threeparttable}
        \end{minipage}
        \begin{minipage}[t]{0.3\textwidth}
            \includegraphics[width=0.9\linewidth]{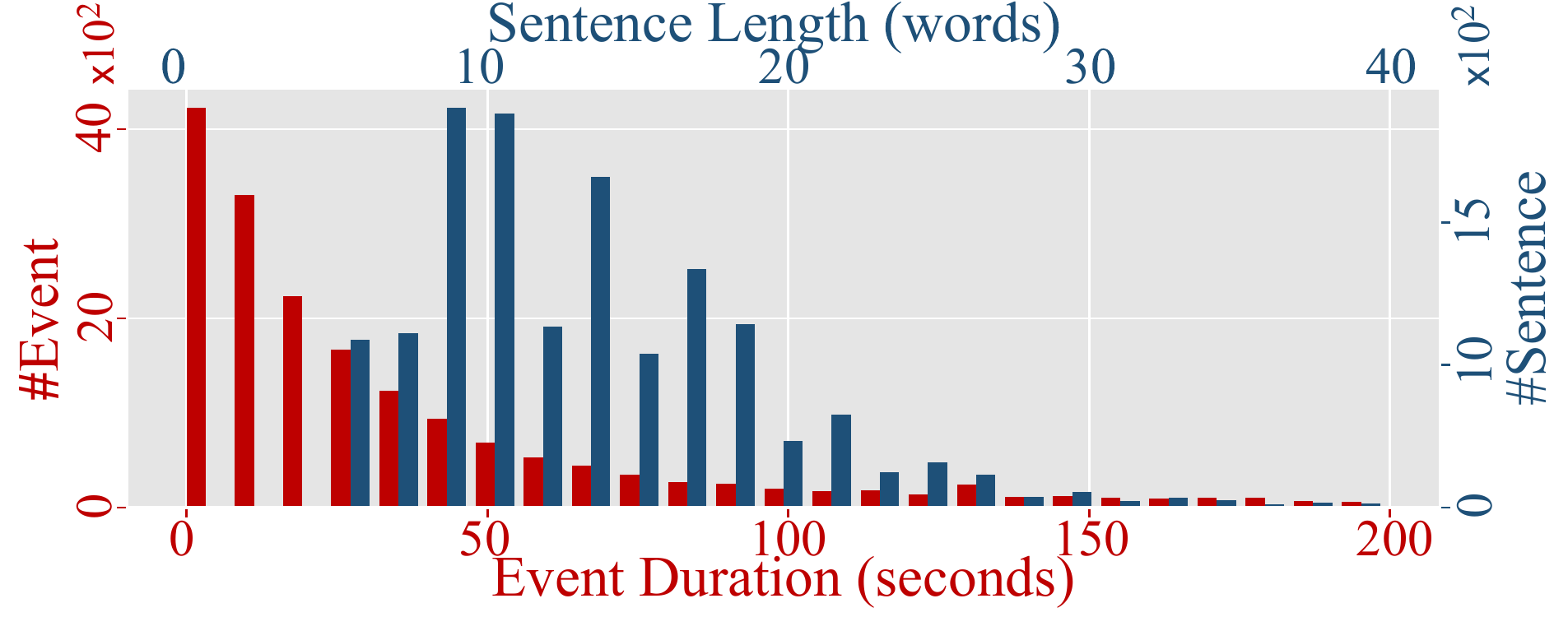}
        \end{minipage}
    \end{minipage}
    \vspace{-6pt}
    \captionlistentry[figure]{A placeholder to increase table counter.}\label{fig:statistics}
    \makeatletter\def\@captype{table}\makeatother\captionsetup{font=small}
    \captionsetup{labelformat=andfigure}
    \caption{Summative statistics of {\DATASET}  dataset (\S\ref{sec:dfs}).}
    \label{tab:statistics}
    \vspace{-17pt}
\end{figure}

\paragraph{Large-Scale.} As shown in Table~\ref{tab:statistics}, {\DATASET} consists of {\cnum{8606}} data examples, collected from {\cnum{3718}} unique videos that span over \cnum{153} hours in total.
On average, each video in {\DATASET} contains \cnum{4.17} events that last \cnum{37.8} seconds, resulting in a total of 15K corresponding descriptive sentences of \cnum{13.5} words.

\paragraph{Dataset Split.}
We separate the {\DATASET} dataset into \texttt{train}/ \texttt{val}/\texttt{test} sets and arrive at a unique split of {\cnum{7053}}/{\cnum{460}}/ {\cnum{1093}} examples
with no overlapping video between \texttt{val}/ \texttt{test} and \texttt{train} sets.
We provide more detailed statistics in both Table~\ref{tab:statistics}\&Fig.~\ref{fig:statistics} and the supplement.

\section{Methodology}\label{sec:md}

\paragraph{Problem$_{\!}$ Statement.$_{\!\!}$}
Given$_{\!}$ a$_{\!}$ video$_{\!}$ $\mathcal{V}$$_{\!}$ with$_{\!}$ $N$ temporally ordered events, \ie, $\mathcal{V}\!=\!\{O_1, \cdots_{\!}, O_{n-1}, H, O_n, \cdots_{\!}, O_{N-1}\}$, the \textit{premise} events, \ie, $\{O_n\}_{n=1}^{N-1}$, and \textit{explanation} event, \ie, $H$, are logically related. The AI system is only presented with a partially observable version of $\mathcal{V}$, \ie, $\tilde{\mathcal{V}}\!=\!\{O_1, \cdots_{\!}, O_{n-1}, \tilde{H}, O_n, \cdots_{\!}, O_{N-1}\}$, where $\tilde{H}$ is obtained by setting all the pixel values of $H$ as zero. The AI system is required to not only describe the premise, but also reason about the most likely explanation for the premise, \ie, gen- erate$_{\!}$ $N$ sentences$_{\!}$ $\mathcal{S}_{\!}\!=_{\!}\!\{S^{O}_n\}^{N-1}_{n=1}\!\cup\!S^{H\!}$ that$_{\!}$ describe$_{\!}$ the$_{\!}$ con- tent of the $N$ events in $\mathcal{V}$, while conditioning on $\tilde{\mathcal{V}}$ only:
\vspace{-4pt}
\begin{equation}\small
\begin{aligned}\label{eq:probability}
    P(\mathcal{S}|\tilde{\mathcal{V}})&=P(S^{H}|\tilde{\mathcal{V}})\prod\nolimits_{n}\!P(S^{O\!}_n|\tilde{\mathcal{V}})\\
    &=\prod\nolimits_{l}\!P(w^{H}_l|w^{H}_{\textless l}, \tilde{\mathcal{V}})\prod\nolimits_{n}\!\prod\nolimits_{l}\!P(w^{On}_l|w^{On}_{\textless l}, \tilde{\mathcal{V}});
\end{aligned}
\vspace{2pt}
\end{equation}
where $w_l$ is the $l$-th word in a generated sentence $S\!\in\!\mathcal{S}$. It is worth mentioning that, when ${H}\!=\!\varnothing$, our \TASKAc~task is degraded into a classic DVC task$_{\!}$~\cite{krishna2017dense}  which focuses only on describing the content of observed events $\{O_n\}_{n=1}^{N-1}$.

\begin{figure*}[t]
    \vspace{-6pt}
    \begin{center}
        \includegraphics[width=1.\linewidth]{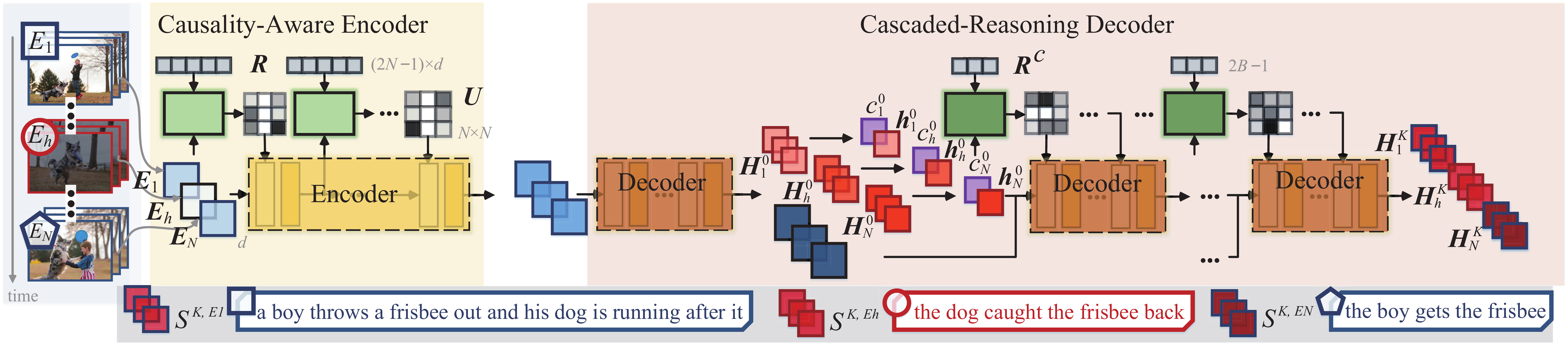}
        \put(-359, 99.5){\scriptsize (\S\ref{sec:ec})}
        \put(-123, 99.5){\scriptsize (\S\ref{sec:dec})}
        \put(-442.5, 71.5){\scriptsize$\Func{Rel}$}
        \put(-401.5, 71.5){\scriptsize$\Func{Rel}$}
        \put(-343.5, 49.5){\scriptsize$\tilde{\Vbm}_1$}
        \put(-338, 39.5){\scriptsize$\tilde{\Vbm}_h$}
        \put(-330.5, 31){\scriptsize$\tilde{\Vbm}_N$}
        \put(-290, 40){\scriptsize$\Dcal^0$}
        \put(-152, 40){\scriptsize$\Dcal^1$}
        \put(-82, 40){\scriptsize$\Dcal^K$}
        \put(-259.5, 35){\tiny$c^0_1$}
        \put(-255, 27){\tiny$c^0_h$}
        \put(-249, 21){\tiny$c^0_N$}
        \put(-194, 71.5){\scriptsize$\Func{Rel}^c$}
        \put(-124, 71.5){\scriptsize$\Func{Rel}^c$}
    \end{center}
    \vspace{-18pt}
    \captionsetup{font=small}
    \caption{\small \textbf{Network architecture} of \OURS. See \S\ref{sec:md} for more details. }
    \label{fig:model}
    \vspace{-12pt}
\end{figure*}

\begin{figure}[t]
    \vspace{-4pt}
    \begin{center}
        \includegraphics[width=1.\linewidth]{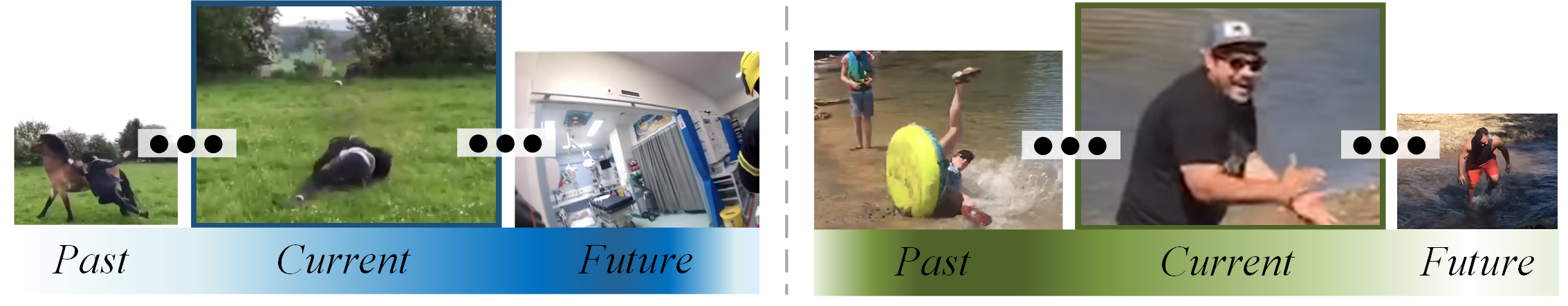}
    \end{center}
    \vspace{-16pt}
    \captionsetup{font=small}
    \caption{\small Illustration of our \textbf{contextualized directional position
embedding} $\Ubm$ (\S\ref{sec:ec}). Darker color indicates larger attention.}
    \label{fig:temporal}
    \vspace{-14pt}
\end{figure}

\paragraph{Core Idea.} Building upon a Transformer encoder-decoder architecture (Fig.~\ref{fig:model}), our {\OURS} is aware of two core challenges posed by the {\TASKAc} task: \textbf{i)} inferring cause-effect relations, and  \textbf{ii)} reasoning beyond the partial$_{\!}$ observation.$_{\!}$
To$_{\!}$ address$_{\!}$ \textbf{i)},$_{\!}$ a$_{\!}$ \textit{contextualized} \textit{directional$_{\!}$ position$_{\!}$ embedding}$_{\!}$ strategy$_{\!}$ is adopted to$_{\!}$ capture$_{\!}$ causal relations residing in the input video $\tilde{\mathcal{V}}$, leading to a \textit{Causality-aware encoder} (\S\ref{sec:ec}). To accommodate \textbf{ii)},  a \textit{confidence-guided multi-step} \textit{reasoning} strategy is developed, \ie, utilize the prediction scores of sentences to guide cross-sentence information flow, yielding a \textit{cascaded-reasoning decoder} (\S\ref{sec:dec}).

\subsection{Causality-Aware Encoder}\label{sec:ec}
For notational simplicity, we redefine the partially obser- vable$_{\!}$ video$_{\!}$ $\tilde{\mathcal{V}}_{\!}\!=_{\!}\!\{O_1, \cdots_{\!}, O_{n-1}, \tilde{H}, O_n, \cdots_{\!}, O_{N-1}\}$$_{\!}$ as$_{\!}$ $\tilde{\mathcal{V}}\!=$ $\{E_n\}_{n=1}^{N}$, where $E_h$ refers to the masked explanation event $\tilde{H}$, and $\{E_n\}_{\neq h\!}$ indicates the visible, premise events $\{O_n\}_{n=1}^{N-1\!}$.{\ms}Let$_{\!}$ us$_{\!}$ denote$_{\!}$ the$_{\!}$ initial$_{\!}$ features$_{\!}$ of$_{\!}$ the$_{\!}$ $N${\ms}events$_{\!}$~as$_{\!}$ $\{\Ebm_{n\!}\!\in$ $\mathbb{R}^{d}\}_{n=1}^{N}$. For each premise event $E_{n\neq h}$, corresponding feature $\Ebm_{n\neq h}$ is obtained by aggregating the visual features of its frames. For the masked explanation event $E_h$, we set $\Ebm_h\!=\!\bm{0}^d$. The Causality-aware encoder is to leverage the context from the  past and/or future observable events $\{E_n\}_{\neq h\!}$ to reinforce their own representations as well as posit a meaningful representation for the most likely explanatory hypothesis, \ie, the masked explanation event $E_h$.

The attention operation is the core of Transformer:
\vspace{-2pt}
\begin{equation}\small\label{eq:attn}
	\!\!\bm{A}\! \sim \bm{X}\bm{W}^{q}(\bm{X}\bm{W}^{k})^{\top\!}, ~~\bm{Y} = \texttt{softmax}(\bm{A})\bm{X}\bm{W}^{v}.
	\vspace{-2pt}
\end{equation}
where the output $\bm{Y}_{\!}\!\in\!\mathbb{R}^{N\times d\!}$ is with the same length $N$ and embedding dimension $d$ as the input $\bm{X}_{\!}\!\in\!\mathbb{R}^{N\times_{\!}d\!}$, and $\bm{W}^{q,k,v\!}\!\in_{\!}\!\mathbb{R}^{d_{\!}\times_{\!}d}$ project$_{\!}$ the$_{\!}$ input into \textit{query}, \textit{key}, and \textit{value} matrices, respectively. As the attention computation is invariant with respect to reordering of the inputs, explicit position encoding is widely adopted, in two typical ways:
\textbf{i)} \textit{Absolute position encoding}$_{\!}$~\cite{vaswani2017attention}: each position $n$ is assigned an embedding, \ie, $\Ubm_{n\!}\!=_{\!}\!\Func{Abs}(n)\!\in\!\RR^{1\x d}$, and the position embeddings are directly added to the input, \ie, $\Xbm\!\leftarrow\!\Xbm\!+\!\Ubm$. $\Func{Abs}(\cdot)$ can be a$_{\!}$ linear$_{\!}$ projection$_{\!}$~\cite{dosovitskiy2020image}, a$_{\!}$ sinusoidal$_{\!}$ function \cite{vaswani2017attention}, \etc. \textbf{ii)} \textit{Relative position encoding} \cite{shaw2018self}: the position
embeddings are constructed considering the pairwise relationships between positions, \ie, $\Ubm_{nm}\!=\!\Func{Rel}(n,m)\!\in\!\RR$.

\paragraph{Contextualized Directional Position Embedding.} Since the {\TASKAc} task is essentially aware of the plausible chains of cause-effect, the relative ordering of the input events matters. We continue in the vein of relative position encoding \cite{shaw2018self,wu2021rethinking} and adopt a \textit{contextualized} \textit{directional$_{\!}$ position$_{\!}$ embedding}$_{\!}$ strategy, \ie, $\Ubm_{nm}\!=\!\Func{Rel}(n,m,\Xbm_n)\!\in\!\RR$:
\vspace{-3pt}
\begin{equation}\small\label{eq:temp_func}
    \begin{aligned}
        \Func{Rel}(n,m,\Xbm_n)\!&=\!
            \Xbm_n\Rbm^\top_{\ell(n,m)}, \\
            \ell(n,m)\!&=\!n-m+N,
    \end{aligned}
    \vspace{-3pt}
\end{equation}
where$_{\!}$ $\Rbm_{\!}\!\in_{\!}\!\RR^{(2N-1)\x d\!}$ is$_{\!}$ a learnable$_{\!}$ matrix,$_{\!}$ and$_{\!}$ $\ell(\cdot,\cdot)$ is a directional indexing function, \ie, $\ell(n,m)\neq \ell(m,n)$. The directional projection $\Func{Rel}$ is conditioned on the visual context, \ie, $\Xbm_n$, since the causal dependency between events is typically related to specific content, \eg, when we see people are laughing, we tend to look back only a short time into the past to figure out the reason; when we see a man falls off his horse, we worry about whether he gets hurt and the impact on his future life. Some more visual examples regarding our contextualized directional position
embedding strategy can be found in Fig.~\ref{fig:temporal}. Then, $\Ubm_{\!}\!\in_{\!}\!\RR^{N_{\!}\x_{\!} N\!}$ is injected by manipulating on the attention matrix$_{\!}$ $\Abm_{\!}\!\in_{\!}\!\RR^{N_{\!}\x_{\!} N_{\!}}$:
\vspace{-3pt}
\begin{equation}\small\label{eq:relative}
    \begin{aligned}
        \Abm_{nm}  \sim \bm{X}_{n}\bm{W}^{q}(\bm{X}_{m}\bm{W}^{k})^{\top\!} + \Ubm_{nm}.
    \end{aligned}
    \vspace{-2pt}
\end{equation}
We further set $\Abm_{nh}\!=\!0$ to encourage leveraging the context from the observable events $\{E_n\}_{\neq h\!}$ to infer the masked explanation event $E_h$,  rather than vice versa. The Causality-aware encoder in  {\OURS} is therefore achieved by stacking several Transformer encoder blocks~\cite{vaswani2017attention} with our contextualized directional position embedding strategy. We denote the output event representations as $\{\tilde{\Vbm}_n\!\in\!\RR^d\}_{n=1}^{N}$.

\subsection{Cascaded-Reasoning Decoder}\label{sec:dec}
With the discriminative representations $\{\tilde{\Vbm}_n\}_{n=1\!}^{N}$ of the observable premise events $\{O_n\}_{n=1}^{N-1\!}$ as well as the explanatory hypothesis $\tilde{H}$,
the cascaded-reasoning decoder first generates a descriptive sentence for each event/hypothesis individually, and then refines all the sentences in a comprehensive, confidence-guided, and step-by-step manner.

\paragraph{Initial$_{\!}$ Description$_{\!}$ Generation.$_{\!}$} For$_{\!}$ each$_{\!}$ event$_{\!}$ representation $\tilde{\Vbm}_n\!\in\!\mathbb{R}^{d}$, a multi-modal, \textit{masked} Transformer decoder is first adopted for initial description generation:
\vspace{-4pt}
\begin{equation}\small\label{eq:dec1}
    \begin{aligned}
        [\tilde{\Vbm}^{0}_n, \Hbm^{0}_n] & = \mathcal{D}^{0}([\tilde{\Vbm}_n, \Hbm_n]),
    \end{aligned}
    \vspace{-4pt}
\end{equation}
where $\Hbm_{n\!}\!\in_{\!}\!\mathbb{R}^{L_{n\!}\x d\!}$ is a set of $L_{n\!}$ words embeddings. During training, it is computed over the groundtruth description, \ie, $\hat{S}^{En}$, and masked attention~\cite{vaswani2017attention} is adopted to prevent the$_{\!~}$ leakage$_{\!~}$ of$_{\!~}$ future$_{\!~}$ words. During$_{\!~}$ inference,$_{\!~}$ it$_{\!~}$ is$_{\!~}$ recur-

\noindent rently generated. Learnable modal-type embeddings$_{\!}$~\cite{devlin2018bert,lei2020mart} are also added into the input yet omitted for brevity.
By fusing visual and linguistic representations as the input, $\mathcal{D}^{0\!}$ conducts cross-modal reasoning, and hence generates improved event representation, \ie, $\tilde{\Vbm}^{0}_n\!\in\!\mathbb{R}^{d}$, and updated visual-linguistic state, \ie, $\Hbm^{0}_n\!\in\!\mathbb{R}^{L_n\x d}$, for each event $E_n$. Then a captioning head is adopted to map $\Hbm^{0}_n$ into word distribution. The probability of $l$-th word is given as:
\vspace{-4pt}
\begin{equation}\small
    \begin{aligned}
        \!\!\!\!\!\!\!\!P(w_l^{En}|w_{<l}^{En\!},\tilde{\mathcal{V}}) &=\! P(w_l^{En}|w_{<l}^{En\!},\Hbm^{0}_n) \\
        &=\! \texttt{softmax}(\Hbm^{0}_n(l)\bm{\Omega}^{\!\top}\!),\!\!
    \end{aligned}
    \vspace{-4pt}
\end{equation}
where $\bm{\Omega}\!\in\!\RR^{|\Omega|\x d\!}$ is the embedding matrix of the word vocabulary $\Omega$, and $\Hbm^{0}_n(l)_{\!}\!\in_{\!}\!\mathbb{R}^{d\!}$ denotes $l$-th vector of $\Hbm^{0}_n$. As standard, the description $S^{0, En\!}\!=_{\!}\!\{w_l^{En}\}_{l=1\!}^{L_n}$ for event $E_{n\!}$ is generated by greedy prediction, and we set the averaged prediction score as the confidence: $c^{0\!}_{n\!}\!=_{\!}\!\frac{1}{L_n\!}\!\sum_i\!P(w_l^{En})$.

\paragraph{Iterative Description Refinement.} To better respond to the fundamental challenge of {\TASKAc} task in reasoning beyond observation, we further cascade several Transformer decoder blocks over $\mathcal{D}^{0\!}$ for iterative description refinement. This allows {\OURS} to make full use of both visual and linguistic context from the past and/or future observable events, and improves the explanatory hypothesis in a step-by-step manner, boosting the reasoning ability eventually.

\begin{figure}[t]
    \vspace{-4pt}
    \begin{center}
        \includegraphics[width=.98\linewidth]{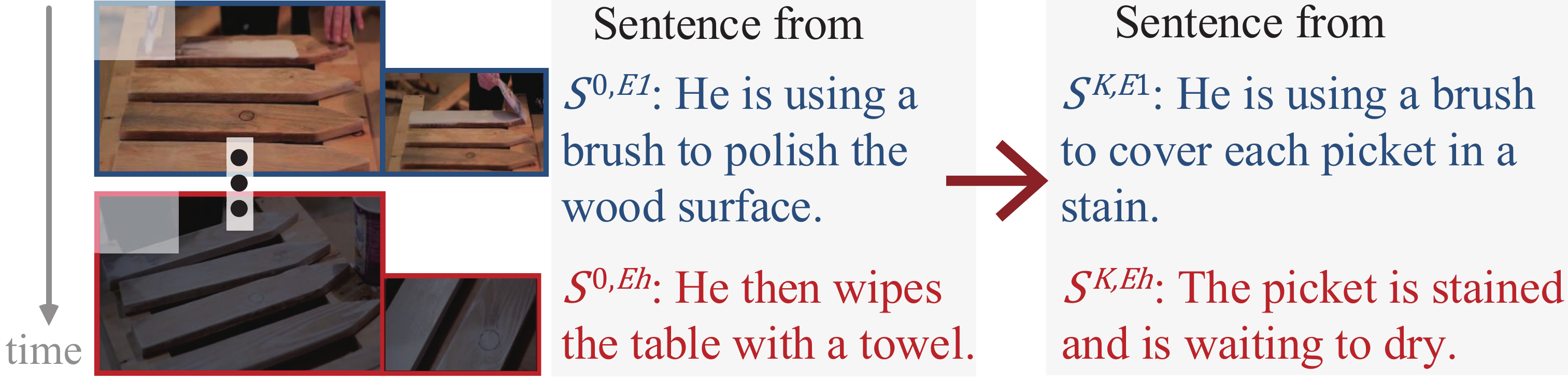}
        \put(-26, 50){\scriptsize$\Dcal^K$}
        \put(-102, 50){\scriptsize$\Dcal^0$}
        \put(-218, 49){\small$E_1$}
        \put(-218, 21){\small$E_h$}
    \end{center}
    \vspace{-18pt}
    \captionsetup{font=small}
    \caption{\small{\m}Sentences from the \textbf{cascaded-reasoning decoder} (\S\ref{sec:dec}).}
    \label{fig:crd}
    \vspace{-14pt}
\end{figure}
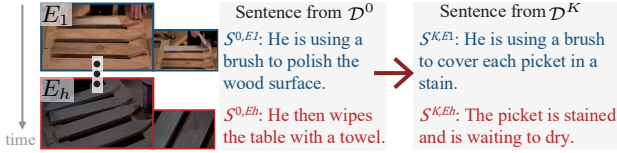

Specifically, our whole refinement procedure can be defined in a recursive, confidence-guided form:
\vspace{-3pt}
\begin{equation}\small\label{eq:dec2}
    \begin{aligned}
        \!\!\!\!\!\![\tilde{\Vbm}^{k}_n, \Hbm^{k}_n]\!=\!\mathcal{D}^{k}([\tilde{\Vbm}^{k-1}_n, &\Hbm_n, \{{\hbm}^{k-1}_n\}^N_{n=1}]), \!~k\!=\!\{1,2,\cdots\!,K\}\!\!\\
        \!\!\!\!\!\!\!\!\!\!\!\!P(w_l^{En}|w_{<l}^{En\!},\tilde{\mathcal{V}}) &\!=\! P(w_l^{En}|w_{<l}^{En\!},\Hbm^{k}_n) \\
                                                                               &\!=\! \texttt{softmax}(\Hbm^{k}_n(l)\bm{\Omega}^{\!\top}\!),\!\!
    \end{aligned}
    \vspace{-4pt}
\end{equation}
where $\mathcal{D}^{k\!}$ refers to $k$-th refinement module and all the refinement modules are weight-sharing Transformer decoders; ${\hbm}^{k-1\!}_n\!\in\!\mathbb{R}^{d}$ indicates a condensed representation of $\Hbm^{k-1\!}_n\!\in\!\mathbb{R}^{L_n\x d}$: ${\hbm}^{k-1\!}_n\!=\!\texttt{maxpool}(\Hbm^{k-1}_n)$. In this way, each $\mathcal{D}^{k\!}$ can leverage inter-sentential relationship in previously generated descriptions $\{{\hbm}^{k-1\!}_n\}^N_{n=1\!}$ for refinement and better reason about the explanatory hypothesis. Moreover, we introduce the event confidence, \ie, $\{c^{k}_n\}^N_{n=1}$,  as a kind of bias into the refinement procedure: leverage the information from those more confident descriptions to help improve the predictions with relatively lower confidence. Without causing ambiguity, we denote $\Xbm$
as the input of the decoder $\mathcal{D}^{k}$, \ie, $\Xbm\!=\![\tilde{\Vbm}^{k-1}_n, \Hbm_n, \{\hbm^{k-1}_n\}^N_{n=1}]$ and omit the superscript $k$. For each input ``token'' $\Xbm_i$, its confidence score $c_{n_i}$ is the one of its sourced event $E_{n_i}$, and we normalize $\{c_n\}^N_{n=1\!}$ over all the $N$ events. Analogous to Eq.$_{\!}$~\ref{eq:relative}, the attention computation in $\mathcal{D}^{k\!}$ is modified as:
\vspace{-3pt}
\begin{equation}\small\label{eq:relative2}
    \begin{aligned}
        &\Abm_{ij}\sim \bm{X}_{i}\bm{W}^{q}(\bm{X}_{j}\bm{W}^{k})^{\top\!} + \Func{Rel}^c(c_{n_i},c_{n_j}),\\
        &\Func{Rel}^c(c_{n_i},c_{n_j})= \rbm^c_{\iota(c_{n_i},c_{n_j})},
    \end{aligned}
    \vspace{-4pt}
\end{equation}
where the learnable vector $\rbm^{c\!}\!\in_{\!}\!\RR^{2B-1\!}$ can be viewed as a bucket to store the relative confidence weight; and the directional indexing function $\iota(\cdot,\cdot)$ is given as $\iota(c_{n_i\!},c_{n_j\!})\!=_{\!}\!\lceil{c_{n_i\!}\!\cdot_{\!}\!B}\rceil\!-\!\lceil{c_{n_j\!}\!\cdot_{\!}\!B}\rceil\!+\!B$.  With such confidence-guided decoding scheme, descriptions are refined by intelligently gathering context from more reliable sentences, while ignoring noisy cues from less confident ones. By stacking several such decoders $\{\mathcal{D}^{k}\}_k$, outputs will be progressively improved (Fig.\!~\ref{fig:crd}). Related experiments can be found in \S\ref{sec:exp:ablation}.

\subsection{Training Objective} \label{sec:train}
\vspace{-1pt}
Given the groundtruth sentences $\{\hat{S}^{En}\}_{n=1\!}^{N\!}$ corresponding to the $N$ events $\{E_n\}_{n=1\!}^{N}$ of video $\tilde{\mathcal{V}}$,
{\OURS} is trained by minimizing the negative log-likelihood over the outputs of the cascaded-reasoning decoder $\{\mathcal{D}^{k\!}\}_{k=0}^K$:
\vspace{-2pt}
\begin{equation}\small
    \Lcal_{\textrm{Main}} = -\textstyle\sum_{k=0}^{K}\textstyle\sum_{n=1}^{N}\textstyle\sum_{l=1}^{L_n} P(\hat{w}_l^{En}|\hat{w}_{<l}^{En\!},\Hbm^{k}_n),
    \vspace{-0pt}
\end{equation}
where$_{\!}$ $\hat{S}^{En\!}\!=_{\!}\!\{\hat{w}_l^{En}\}_{l=1}^{L_n}$.$_{\!}$ As$_{\!}$ the$_{\!}$ teacher$_{\!}$ forcing$_{\!}$ scheme$_{\!}$~\cite{williams1989learning}$_{\!}$ is used for training, $\Hbm_n$ in Eq.~\ref{eq:dec1} and \ref{eq:dec2} is embedded over one-hot encoded groundturth words $\{\hat{w}_l^{En}\}_{l}$. We further adopt a \textit{hypothesis reconstruction} based optimization criterion, to provide the encoder with more explicit supervision signals for explanatory hypothesis reasoning:
\vspace{-2.5pt}
\begin{equation}\small \label{eq:aux}
    \Lcal_{\textrm{Aux}} = \norms{\Func{Proj}(\tilde{\bm{V}}_h) - \Func{Proj}(\hat{\bm{V}}_h)}_2,
    \vspace{-3pt}
\end{equation}
where $\tilde{\bm{V}}_{h\!}$ and $\hat{\bm{V}}_{h\!}$ are embeddings for the explanatory hypothesis obtained from the masked and original videos, \ie, $\tilde{\mathcal{V}}$ and ${\mathcal{V}}$, respectively, and $\Func{Proj\!}$ is a projection head, based on a small multi-layer perceptron.  This auxiliary training objective forces {\OURS} to ``imagine'' an effective representation $\tilde{\bm{V}}_{h}$ that better aligns with the original content of $E_{h}$. $\hat{\bm{V}}_{h\!}$ is from the momentum version of the encoder.

\subsection{Implementation Details}
\vspace{-1pt}
Details on implementing the algorithm are as follows:
\begin{fullitemize}
    \item \textit{Detailed network architecture}: The encoder (\S\ref{sec:ec}) of {\OURS} is implemented as two Transformer encoder blocks, and each decoder module (\S\ref{sec:dec}), \ie, $\mathcal{D}^{k}$, is implemented as two Transformer masked decoder blocks. They have $d\!=\!768$ hidden size and \cnum{12} attention heads. We use a bucket size $B\!=\!10$ to quantize confidence scores (Eq.\!~\ref{eq:relative2}). We stack a total of $K\!=\!3$ decoders for cascaded reasoning.
    \item \textit{Data preprocessing}: For each video event, action/appearance features are pre-extracted using ActivityNet\tcite{caba2015activitynet} pre-trained ResNet200\tcite{he2016deep}/BN-Inception\tcite{ioffe2015batch}, as in\tcite{zhou2018end,lei2020mart,wang2021end}. We uniformly sample \cnum{50} frames per event and concatenate their features as the corresponding event representation which is denoted in a vector form in \S\ref{sec:ec}-\ref{sec:train} for ease of notation. Sentences are padded or truncated into \cnum{20} words.
    \item \textit{Training/Inference}: For the first decoder $\mathcal{D}^{0}$, we adopt scheduled sampling~\cite{bengio2015scheduled} to make the later decoders fully trained. The coefficient between the main and auxiliary training objectives is set as \cnum{0.2}. During inference, the final descriptive sentences are generated from the last decoder $\mathcal{D}^{K}$, using  deterministic decoding, \ie, greedy search. All the experiments are conducted on \cnum{2} NVIDIA GeForce RTX 2080 Ti GPUs with a \cnum{11}GB memory per-card.
\end{fullitemize}

\begin{table*}[t]
    \centering\small
    \resizebox{1.\textwidth}{!}{
        \setlength\tabcolsep{2pt}
        \renewcommand\arraystretch{1.03}
        \begin{tabular}{x{74}||cc|ccccc|ccccc}
            \hline\thickhline
            \rowcolor{mygray}
                                                                           & &  & \multicolumn{5}{c|}{\textbf{Premise Event}} & \multicolumn{5}{c}{\textbf{Explanation Event}} \\
            \rowcolor{mygray}
            \multirow{-2}{*}{Method} &\multirow{-2}{*}{Encoder} & \multirow{-2}{*}{Decoder} & BLEU@4 & METEOR & ROUGE & CIDEr & BERT-S & BLEU@4 & METEOR & ROUGE & CIDEr & BERT-S \\
            \hline \hline
            Human                                                          & -     & -       & {\void{20}{10}{13.26}}& {\void{20}{10}{21.27}}& {\void{20}{10}{39.47}}& {\void{20}{10}{155.72}} & {\void{20}{10}{45.33}} & {\void{20}{10}{11.35}} & {\void{20}{10}{19.36}}& {\void{20}{10}{36.92}}& {\void{20}{10}{147.79}} & {\void{20}{10}{40.59}}\\
            \hline
            \subt{48}{22}{VTrans  \cite{zhou2018end}}{\sub{CVPR18}}       & Trans. & Trans.  & {\void{20}{10}{4.20}}& {\void{20}{10}{9.94}} & {\void{20}{10}{21.13}} & {\void{20}{10}{31.09}} & {\void{20}{10}{29.05}} & {\void{20}{10}{0.71}}& {\void{20}{10}{6.92}}& {\void{20}{10}{19.12}} & {\void{20}{10}{7.11}} & {\void{20}{10}{22.13}} \\
            \subt{48}{22}{MFT     \cite{xiong2018move}}{\sub{ECCV18}}     & RNN    & RNN     & {\void{20}{10}{3.93}}& {\void{20}{10}{9.69}} & {\void{20}{10}{20.81}} & {\void{20}{10}{30.96}} & {\void{20}{10}{27.41}} & {\void{20}{10}{1.81}}& {\void{20}{10}{7.16}}& {\void{20}{10}{19.16}} & {\void{20}{10}{17.67}} & {\void{20}{10}{25.90}} \\
            \subt{48}{22}{Trans-XL\cite{dai2019transformer}}{\sub{ACL19}} & Trans. & Trans.  & {\void{20}{10}{3.98}}& {\void{20}{10}{9.53}} & {\void{20}{10}{21.02}} & {\void{20}{10}{30.87}} & {\void{20}{10}{29.12}} & {\void{20}{10}{2.96}}& {\void{20}{10}{7.51}}& {\void{20}{10}{20.94}} & {\void{20}{10}{24.54}} & {\void{20}{10}{27.23}} \\
            \subt{48}{22}{MART    \cite{lei2020mart}}{\sub{ACL20}}        & Trans. & Trans.  & {\void{20}{10}{3.74}}& {\void{20}{10}{9.48}} & {\void{20}{10}{21.17}} & {\void{20}{10}{29.22}} & {\void{20}{10}{29.03}} & {\void{20}{10}{2.86}}& {\void{20}{10}{7.47}}& {\void{20}{10}{20.87}} & {\void{20}{10}{24.05}} & {\void{20}{10}{27.77}} \\
            \subt{48}{22}{PDVC    \cite{wang2021end}}{\sub{ICCV21}}       & Trans. & RNN     & {\void{20}{10}{4.28}}& {\void{20}{10}{9.95}} & {\void{20}{10}{21.19}} & {\void{20}{10}{33.59}} & {\void{20}{10}{29.37}} & {\void{20}{10}{3.00}}& {\void{20}{10}{8.54}}& {\void{20}{10}{20.71}} & {\void{20}{10}{25.14}} & {\void{20}{10}{27.80}} \\\hline
            \textbf{{{\OURS}}}                                    & Trans. & Trans.  & {\bbetter{20}{10}{\textbf{5.03}}{0.72}}& {\bbetter{20}{10}{\textbf{10.75}}{0.80}} & {\bbetter{20}{10}{\textbf{24.81}}{3.62}} & {\bbetter{20}{10}{\textbf{38.27}}{4.68}} & {\bbetter{20}{10}{\textbf{34.88}}{5.51}} & {\bbetter{20}{10}{\textbf{3.44}}{0.44}}& {\bbetter{20}{10}{\textbf{9.05}}{0.49}}& {\bbetter{20}{10}{\textbf{22.89}}{1.95}} & {\bbetter{20}{10}{\textbf{30.75}}{5.61}} & {\bbetter{20}{10}{\textbf{30.64}}{2.84}} \\
            \hline
        \end{tabular}
    }
    \captionsetup{font=small}
    \caption{\small \textbf{Quantitative results} on the  \texttt{test} set of our {\DATASET} dataset. `Trans.' indicates Transformer-based architecture. See \S\ref{sec:exp:main} for details.}
    \label{tab:vad}
    \vspace{-8pt}
\end{table*}

\begin{figure*}
    \begin{minipage}{\textwidth}
        \begin{minipage}[t]{0.65\textwidth}
            \begin{center}
                \includegraphics[width=1.\linewidth]{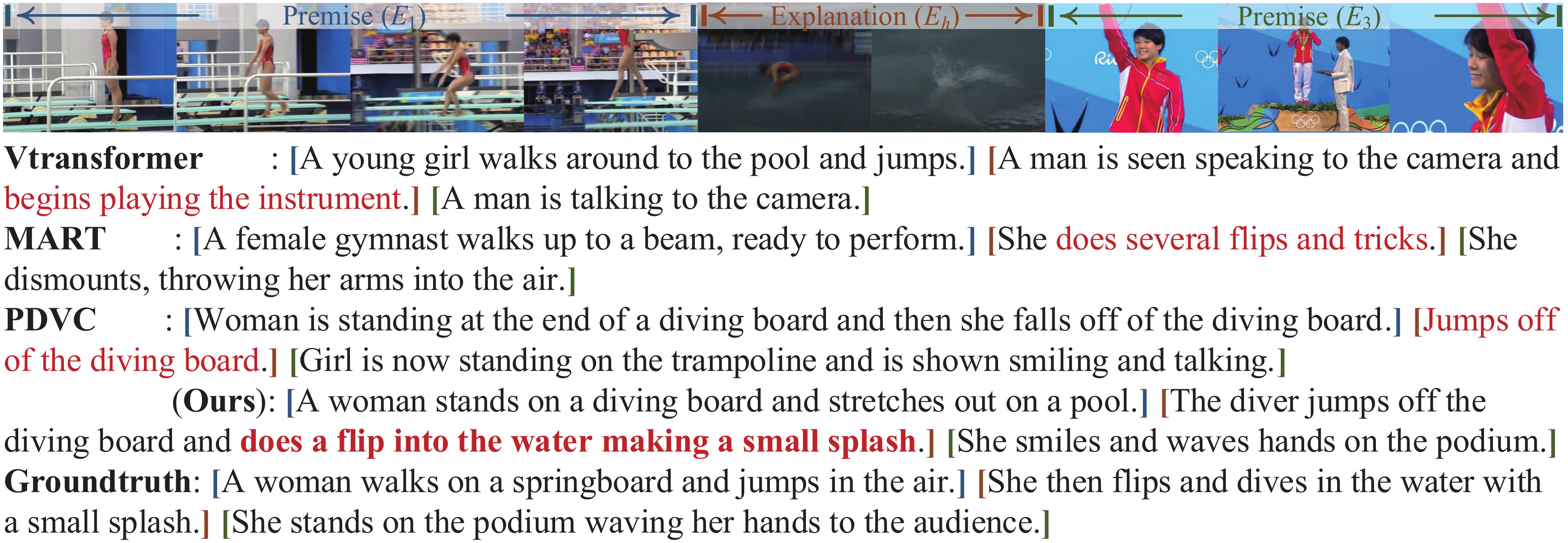}
                \put(-323, 27){\scriptsize \textbf{\OURS}}
                \put(-302.5, 44){\scriptsize \cite{wang2021end}}
                \put(-300.5, 60.5){\scriptsize \cite{lei2020mart}}
                \put(-280.5, 77.5){\scriptsize \cite{zhou2018end}}
            \end{center}
            \vspace{-18pt}
            \figurecaption{Qualitative comparison}
            {(\S\ref{sec:exp:main}) of {\OURS} and \cite{wang2021end,lei2020mart,zhou2018end} on {\DATASET} \texttt{test}.}
            {qualitative}
        \end{minipage}
        \hspace{0.5ex}
        \begin{minipage}[t]{0.33\textwidth}
            \vspace{-1.53in}
            \centering\small
            \resizebox{1.\linewidth}{!}{
                \setlength\tabcolsep{8pt}
                \renewcommand\arraystretch{1.}
                \begin{tabular}{r|c|l}
                    \hline\thickhline
                    \rowcolor{mygray}
                    \multicolumn{3}{c}{\textbf{Premise Event}}    \\
                    \hline
                    \multicolumn{1}{c|}{Prefer A} & Neutral & \multicolumn{1}{c}{Prefer B}                 \\
                    \textbf{{\OURS}}~~\textbf{34.2}       & 41.4   & 15.9~~PDVC~\cite{wang2021end} \\
                    \textbf{{\OURS}}~~16.0       & 35.3   & \textbf{39.5}~~Human                   \\
                    \hline\thickhline
                    \rowcolor{mygray}
                    \multicolumn{3}{c}{\textbf{Explanation Event}}    \\
                    \hline
                    \multicolumn{1}{c|}{Prefer A} & Neutral & \multicolumn{1}{c}{Prefer B}  \\
                    \textbf{{\OURS}}~~\textbf{29.9}       & 13.7   & 10.4~~PDVC~\cite{wang2021end} \\
                    \textbf{{\OURS}}~~~~8.9       & 22.1   & \textbf{64.8}~~Human                   \\
                    \hline
                \end{tabular}
            }
            \vspace{-6pt}
            \captionsetup{font=small}
            \makeatletter\def\@captype{table}\makeatother\captionsetup{font=small}
            \caption{\small \textbf{User study} of pairwise model preference (\%). ``Neutral'' means A and B models are ``equally
        good''. Percentage of ``equally bad'' are omitted. See \S\ref{sec:exp:main} for details.}
            \label{tab:User_study}
        \end{minipage}
    \end{minipage}
    \vspace*{-12pt}
\end{figure*}

\vspace{-4pt}
\section{Experiments}\label{sec:exp}
\vspace{-1pt}
$_{\!\!}$We$_{\!}$ first$_{\!}$ provide$_{\!}$ benchmarking$_{\!}$ results$_{\!}$ on$_{\!}$ our$_{\!}$ {\DATASET}$_{\!}$ dataset$_{\!}$ (\S\ref{sec:exp:main}). Then,  to verify the efficacy of our core model designs, we  conduct  a set of diagnostic studies (\S\ref{sec:exp:ablation}). Finally, for comprehensive evaluation, we test our  {\OURS} on the classic, dense video captioning (DVC) task~\cite{krishna2017dense} (\S\ref{sec:exp:dvc}).

\vspace{-1pt}
\subsection{Performance on {\DATASET} Task}\label{sec:exp:main}
\vspace{-1pt}

\paragraph{Competitor.}\label{sec:baseline}
We benchmark five{\ms}top-leading{\ms}DVC{\ms}models{\ms}on{\ms}{\TASKAc}{\ms}to{\ms}reveal{\ms}the{\ms}abductive{\ms}reasoning{\ms}ability{\ms}in{\ms}existing{\ms}techniques.{\ms}
They{\ms}include three Transformer-based~\cite{zhou2018end,dai2019transformer,lei2020mart} and two RNN-based~\cite{xiong2018move,wang2021end} models, which are trained on \texttt{train} set of our VAR dataset with pre-provided event segments using their original training protocols.

\paragraph{Evaluation Metric.} Five well-known automated metrics, \ie, BLEU@4~\cite{papineni2002bleu}, CIDEr~\cite{vedantam2015cider}, METEOR~\cite{banerjee2005meteor}, ROUGE-L~\cite{lin2002manual}, and BERTScore~\cite{zhang2019bertscore}, are used for evaluation.

\def\ADDINTEXT{
\vspace{-2.5ex}
\subsection{Diagnostic Experiment}\label{sec:exp:ablation}
\vspace{-0.5ex}
A set of ablative studies is conducted on {\DATASET} \texttt{test} for indepth analyzing each component in$_{\!}$ our$_{\!}$ {\OURS},$_{\!}$ using$_{\!}$ BLEU@4,$_{\!}$ CIDEr$_{\!}$~and BERT-S metrics, averaged over all the events.

\paragraph{Key Component Analysis.} We first study the efficacy of core model components. The first row in Table$_{\!}$~\ref{tab:AS1} gives the performance of a basic Transformer model, which simply uses ab-
}

\begin{figure*}[t!]
    \begin{minipage}{\textwidth}
        \vspace{-.75ex}
        {
            \begin{minipage}[t]{0.375\textwidth}
                \begin{minipage}[t][0.1255\textheight][t]{0.5\textwidth}
                    ~
                \end{minipage}\\
                {\normalsize
                \ADDINTEXT
                }
            \end{minipage}
        }
        \hspace{.05in}
        \begin{minipage}[t]{0.62\textwidth}
            \vspace{-2ex}
            \begin{threeparttable}[t]
                {\hspace{-0.645\textwidth}
                \subfloat[\scalebox{.95}{Key components} \label{tab:AS1}]{
                    \grouptablestyle{3pt}{1}
                        \begin{tabular}{c|cc||ccc}
                            \hline\thickhline
                            \rowcolor{mygray}
                                                & Causality-Aware   & Cascaded-Reasoning         &        &       &   \\
                            \rowcolor{mygray}
                            \multirow{-2}{*}{\#} & {Encoder (\S\ref{sec:ec})} & {Decoder (\S\ref{sec:dec})}   & \multirow{-2}{*}{BLEU@4} & \multirow{-2}{*}{CIDEr} & \multirow{-2}{*}{BERT-S} \\
                            \hline \hline
                            1                    &                    &                  & 3.39 & 30.04 & 26.35 \\
                            2                    & \ding{51}          &                  & 3.91 & 32.32 & 29.85 \\
                            3                    &                    & \ding{51}        & 4.05 & 33.71 & 29.94 \\
                            4                    & \ding{51}          & \ding{51}        & 4.66 & 36.13 & 33.44 \\
                            \hline
                        \end{tabular}
                }
                }
                \hspace{.6ex}
                \parbox{.15\textwidth}{
                    \vspace{0.03in}
                    \subfloat[\scalebox{.95}{Position embedding strategy}\label{tab:AS2}]{%
                    \grouptablestyle{3pt}{1.}
                        \begin{tabular}{cc||ccc}
                            \hline\thickhline
                            \rowcolor{mygray}
                            $\Ubm_{n}$/$\Ubm_{nm\!}$   &          &        &       &   \\
                            \rowcolor{mygray}
                             (\S\ref{sec:ec}) & \multirow{-2}{*}{Formulation}  & \multirow{-2}{*}{BLEU@4}                         & \multirow{-2}{*}{CIDEr} & \multirow{-2}{*}{BERT-S} \\
                            \hline \hline
                            Absolute                    &$\Ubm_{n\!}\!=_{\!}\!\Func{Abs}(n)$                                    & 4.20 & 33.27 & 29.95 \\
                            Directional          & $\Ubm_{nm\!}\!=_{\!}\!\Func{Rel}(n,m)$                                       & 4.35 & 34.25 & 31.79 \\
                            \tabincell{c}{Contextualized\\Directional}& $\Ubm_{nm\!}\!=_{\!}\!\Func{Rel}(n,m,\bm{X}_{\!n\!})$   & 4.66 & 36.13 & 33.44 \\
                            \hline
                        \end{tabular}
                    }

                }
                \vfill\vspace{-.11in}
                \subfloat[\scalebox{.95}{Cascaded reasoning} \label{tab:AS3}]{
                        \grouptablestyle{4pt}{1.11}%
                            \begin{tabular}{x{35}||ccc}
                                \hline\thickhline
                                \rowcolor{mygray}
                                $\mathcal{D}^{k\!}$ (\S\ref{sec:dec})   & {B@4}  & {CIDEr} & {BERT-S} \\
                                \hline \hline
                                $K = 0$                & 3.91 & 32.72 & 29.50 \\
                                $K = 1$                & 4.34 & 34.89 & 31.60 \\
                                $K = 2$                & 4.61 & 35.53 & 32.57 \\
                                $K = 3$                & 4.66 & 36.13 & 33.44 \\
                                $K = 4$                & 4.66 & 36.05 & 33.51 \\
                                $K = 5$                & 4.60 & 35.90 & 33.32 \\
                                \hline
                            \end{tabular}
                }
                \hspace{.6ex}
                \parbox{.2\textwidth}{
                    \vspace{0.03in}
                    \subfloat[\scalebox{.95}{Confidence embedding} \label{tab:AS4}]{
                        {\grouptablestyle{3.5pt}{1.}%
                        \begin{tabular}{x{40}||ccc}
                            \hline\thickhline
                            \rowcolor{mygray}
                            $\Func{Rel}^{c\!}$ (Eq.$_{\!}$~\ref{eq:relative2})  & {BLEU@4} & {CIDEr} & {BERT-S} \\
                            \hline \hline
                                                   & 4.45 & 35.22 & 33.17 \\
                            \ding{51}              & 4.66 & 36.13 & 33.44 \\
                            \hline
                        \end{tabular}
                        }
                    }\\\vspace{-.16in}
                    \subfloat[\scalebox{.95}{Training objective} \label{tab:AS5}]{
                        {\grouptablestyle{3.5pt}{1.}%
                            \begin{tabular}{x{40}||ccc}
                                \hline\thickhline
                                \rowcolor{mygray}
                                Loss (\S\ref{sec:train})  & {BLEU@4}                                      & {CIDEr} & {BERT-S} \\
                                \hline \hline
                                $\mathcal{L}_{\text{Main}}$                                               & 4.40 & 35.51 & 32.83 \\
                                $\mathcal{L}_{\text{Main}}\!\!+\!\!\mathcal{L}_{\text{Aux}}$              & 4.66 & 36.13 & 33.44 \\
                                \hline
                            \end{tabular}
                        }
                    }
                }
                \vfill
                \vspace{-8pt}
                \makeatletter\def\@captype{table}\makeatother\captionsetup{font=small}
                \caption{\small A set of \textbf{ablation studies} (\S\ref{sec:exp:ablation}) on the \texttt{test} set of our VAR dataset.}
                \label{tab:ablations}
            \end{threeparttable}
        \end{minipage}
    \end{minipage}
    \vspace*{-20pt}
\end{figure*}

\paragraph{Quantitative$_{\!}$ Result.$_{\!\!}$} Table$_{\!}$~\ref{tab:vad}$_{\!}$ summarizes$_{\!}$ the$_{\!}$ benchmarking results on the \texttt{test} set of our {\DATASET} dataset. For detailed analysis, we report the performance over the observable premise events and invisible explanation events separately. Moreover, to probe the upper bound of model performance, we evaluate human performance by asking ten volunteers to perform {\TASKAc}. Specifically, we randomly sample \cnum{500} examples from unique videos in {\DATASET} \texttt{test}. The volunteers are only provided with partially observable videos and requested to write down the corresponding descriptions and hypotheses. The human-written descriptions and hypotheses are  evaluated by the automatic metrics, and evaluation scores are shown in the first row of Table~\ref{tab:vad}. Several essential conclusions can be drawn from Table~\ref{tab:vad}: \textbf{i)} Humans are good at {\TASKAc}; although human-written hypotheses for explanation scored lower than the descriptions for the visual premise, they are still very plausible in absolute terms. \textbf{ii)} All traditional DVC models~\cite{xiong2018move,wang2021end,zhou2018end,dai2019transformer,lei2020mart} struggle with {\TASKAc} that humans excel at. Their generated hypotheses are usually untrusted, and far worse than their created premise narratives. This suggests that existing video-based language generation models are not good at reasoning beyond observation.  \textbf{iii)} Our {\OURS} outperforms other AI models~\cite{xiong2018move,wang2021end,zhou2018end,dai2019transformer,lei2020mart}, in both explanatory hypothesis reasoning and premise description, demonstrating the effectiveness of our whole model design. Compared to other AI models, {\OURS} also yields a relatively smaller performance drop, from premise description to hypothesis reasoning. This suggests that {\OURS} can make better use of the context of observed events to infer the explanatory hypothesis. \textbf{iv)} Although our {\OURS} shows more promising results, there still remains a significant gap from human performance, that is waiting for more sophisticated abductive reasoning models to conquer.

\paragraph{User Study.} For comprehensive performance assessment, we further carry out a subjective evaluation, based on pairwise model comparison. Specifically, we randomly sample \cnum{500} examples from unique videos in {\DATASET} \texttt{test}. Three volunteers are presented the outputs of a pair of systems (\ie, {\OURS} \textit{vs} PDVC~\cite{wang2021end} or human) on the sampled examples, and requested to do a comparison about which one is better, or ``equally good'' or ``equally bad''. The human preference results are collected in Table$_{\!}$~\ref{tab:User_study}, and again the statistics for premise events and explanation events are presented separately. The human subjective judgments are generally accordant with the trends reflected by Table~\ref{tab:vad}. Specifically,  the human pairwise comparison results confirm the superiority of {\OURS} over PDVC, the second-best model in Table~\ref{tab:vad}:  {\OURS} receives \cnum{34.2} and \cnum{29.9} percent preference votes on the premise description and explanatory
hypothesis, respectively. However, human-written hypotheses and descriptions are much more favorable than our results,  showing again {\TASKAc} is a very challenging task.

\paragraph{Qualitative Analysis.} A test video example in {\DATASET} dataset is shown in Fig.~\ref{fig:qualitative}. It contains the explanatory hypotheses and premise descriptions from our {\OURS} and {other competitors~\cite{wang2021end,lei2020mart,zhou2018end}} as well as groundtruth sentences. We can find that our {\OURS} is able to discover and correctly describe the cause-effect chain, and hence generate a plausible hypothesis, \ie, \textit{making a small splash}, that well explains the observed events, \ie, \textit{standing on the podium}. In contrast, other competitors typically produce unsatisfactory results, especially for the explanatory hypothesis.

\noindent solute position embedding in the encoder and only adopts one single decoder, \ie, $\mathcal{D}^{0}$.
The  results  in the first two rows reveal that contextualized directional position embedding (\S\ref{sec:ec}) consistently improves the performance over the three metrics.
Moreover, from the first and third rows we can observe that confidence-guided multi-step reasoning (\S\ref{sec:dec}) indeed boosts the performance. By further considering the scores in the last row, we can safely conclude that combining the two model designs together leads to the best results.

\paragraph{Contextualized Directional Position Embedding.} Next, to thoroughly study the impact of our contextualized directional position embedding strategy (\S\ref{sec:ec}), we report the performance of two alternatives in Table$_{\!}$~\ref{tab:AS2}. Specifically, ``absolute'' refers to the widely used, learnable absolute position embedding, while ``directional'' indicates learning relative position embedding without considering any input context. As seen, our contextualized directional position embedding is significantly better than the two alternatives.

\paragraph{Cascaded Reasoning.$_{\!}$} Table~\ref{tab:AS3} reports the performance with different steps of our cascaded reasoning (\S\ref{sec:dec}), \ie,  $K\!=\!\{0, 1, \cdots\!, 5\}$. When $K\!=\!0$, only one decoder $\mathcal{D}^{0}$ is adopted and the CIDEr score is just \cnum{32.72}. However, after adding an extra refinement decoder, the score is greatly improved to \cnum{36.13}. The increasing trend is gradually saturated until $K\!>\!3$. We therefore use  $K\!=\!3$  as our default setting for balancing performance and inference efficiency.

\paragraph{Confidence Embedding.$_{\!}$} We inject sentence scores into~the cascaded reasoning for guiding information flow (Eq.$_{\!}$~\ref{eq:relative2}).~As shown in$_{\!}$ Table$_{\!}$~\ref{tab:AS4}, removing confidence embedding hinders the performance, \eg, \cnum{36.13}$\rightarrow$\cnum{35.22} in terms of CIDEr.

\paragraph{Training Objective.} Finally we examine our training objective design (\S\ref{sec:train}). Table$_{\!}$~\ref{tab:AS5} demonstrates a beneficial impact of the hypothesis reconstruction loss $\mathcal{L}_{\text{Aux}}$ (Eq.\!~\ref{eq:aux}).

\vspace{-2pt}
\subsection{Performance on DVC Task}\label{sec:exp:dvc}
\vspace{-2pt}
For completeness, we report performance on DVC task.
\paragraph{Dataset.}
As a gold-standard dataset for DVC, ActivityNet{\ms}Captions\tcite{krishna2017dense}{\ms}contains{\ms}a{\ms}total{\ms}of{\ms}20k{\ms}untrimmed{\ms}videos (\cnum{10009}/\cnum{4917}/\cnum{5044} for \texttt{train}/\texttt{val}/\texttt{test}). Each video lasts 120s and is annotated with 3.65 temporally-localized sentences on average. Following~\cite{zhou2018end,lei2020mart,song2021towards}, \texttt{val} set is further split into two subsets: \texttt{ae-val} with \cnum{2460} videos and \texttt{ae-test} with \cnum{2457} videos without overlapping.

\paragraph{Evaluation Metric.} As in\tcite{song2021towards,lei2020mart,zhou2018end}, BLEU@4~\cite{papineni2002bleu}, METEOR~\cite{banerjee2005meteor}, and CIDEr~\cite{vedantam2015cider} metrics are used for evaluation.
\paragraph{Quantitative Result.} {\OURS} is trained on the \texttt{train} set and evaluated on \texttt{ae-val} set in paragraph-level. Since we focus only on descriptive quality, the sentences are generated from a provided list of events, like in~\cite{park2019adversarial,lei2020mart,ji2021hierarchical}. As shown in Table~\ref{tab:densecap},  {\OURS} outperforms state-of-the-art DVC models over all the metrics, \eg, $+2.81$ performance gain in CIDEr. This proves the strong reasoning ability of {\OURS} and emphasizes the value of our VAR task in promoting innovations of powerful video-language models.

\begin{table}[t]
    \centering\small
    \resizebox{0.44\textwidth}{!}{
        \hspace{-2ex}
        \setlength\tabcolsep{2pt}
        \renewcommand\arraystretch{1.}
        \begin{tabular}{c||ccc}
            \hline\thickhline
            \rowcolor{mygray}
            Method                                                         & BLEU@4 & METEOR & CIDEr \\
            \hline \hline
            \subt{50}{21}{HSE~\cite{zhang2018cross}}{\sub{ECCV18}}         & {\void{20}{15}{9.84}} & {\void{20}{15}{13.78}} & {\void{20}{15}{18.78}}  \\
            \subt{50}{21}{Trans-XL~\cite{dai2019transformer}}{\sub{ACL19}} & {\void{20}{15}{10.39}} & {\void{20}{15}{15.09}} & {\void{20}{15}{21.67}} \\
            \subt{50}{21}{VTrans~\cite{zhou2018end}}{\sub{CVPR18}}         & {\void{20}{15}{9.75}} & {\void{20}{15}{15.64}} & {\void{20}{15}{22.16}} \\
            \subt{50}{21}{MART~\cite{lei2020mart}}{\sub{ACL20}}            & {\void{20}{15}{10.33}} & {\void{20}{15}{15.68}} & {\void{20}{15}{23.42}} \\
            \subt{50}{21}{PDVC~\cite{wang2021end}}{\sub{ICCV21}}           & {\void{20}{15}{11.80}} & {\void{20}{15}{15.93}} & {\void{20}{15}{27.27}} \\
            \hline
            \textbf{\OURS}                                       & {\bbetter{20}{15}{\textbf{12.45}}{0.65}} & {\bbetter{20}{15}{\textbf{16.43}}{0.50}} & {\bbetter{20}{15}{\textbf{30.08}}{2.81}} \\
            \hline
        \end{tabular}
    }
    \captionsetup{font=small}
    \caption{\small $_{\!\!}$\textbf{Quantitative$_{\!}$ results$_{\!}$} (\S\ref{sec:exp:dvc})$_{\!}$ on$_{\!}$ the$_{\!}$ \texttt{ae-val}$_{\!}$ set$_{\!}$  of$_{\!}$  Activi- tyNet Captions~\cite{krishna2017dense}. The scores are mainly borrowed from~\cite{wang2021end}. }
    \label{tab:densecap}
    \vspace{-12pt}
\end{table}

\vspace{-4pt}
\section{Conclusion}
\vspace{-4pt}
We introduce \TASKAc~({\TASK}) -- a novel task that investigates the abductive reasoning ability of machine intelligence in the visual world. We establish {\OURS}, a new Transformer based visual-language model, which captures the context from visual premise in a causality-aware manner, and generates premise descriptions and hypothesis sentences in a confidence-guided, step-by-step fashion. {\OURS} shows promising results on both \TASKAc~and dense video captioning tasks. We also observe a remaining large headroom for AI systems in \TASKAc, which is expected to  encourage exciting avenues in the future.

\appendix
{

{
\it
	In this work, we build a headway of {\TASK} ({\TASKAc}) as a new task and introduce a large-scale dataset for {\TASKAc} that
	scaffolds the investigation of Abductive Reasoning ability in AI systems.
	Along with them, {\OURS} is further presented as a modest solution.
	In the supplemental material, we provide the following items that shed deeper insight on the aforementioned contributions:
	
	\begin{itemize}[labelwidth=!,itemsep=0pt,topsep=1pt,parsep=1pt]
		\item Additional dataset analysis (\secref{dataset})
		\item Additional details of benchmarked baselines (\secref{baseline})
		\item {Additional experimental results} (\secref{sexp})
		\item Additional qualitative visualization (\secref{qualitative})
		\item Discussion of limitation and reproductibility (\secref{limitation})
		\item Discussion of legal and ethical considerations (\secref{legal})
	\end{itemize}
}

\vspace{2ex}
\section{Additional Dataset Analysis}\label{sec:dataset}

\subsection{Detailed Dataset Statistics} \label{sec:statistic}
Our {\DATASET} dataset is curated from three main sources, \ie, YouTube Lifestyle video, movie and TV show,
in \figref{AS2} and \figref{AS3},
we illustrate the detailed distribution of videos and examples by collected sources.
As seen, most videos in {\DATASET} are from YouTube Lifestyle video, while movie videos tend to have more events, and thus leads to more complicated causal structures and more examples.
We then study the distribution of frequently used words in {\DATASET} descriptions, which is illustrated as a word cloud in \figref{AS1}. More frequent words are shown in larger font size.
Finally, the distribution of premise events is shown in \figref{AS4}.

\begin{figure}[t]
    \subfloat[{Most frequently used words} \lblfig{AS1}]{
        \includegraphics[width=1.\linewidth]{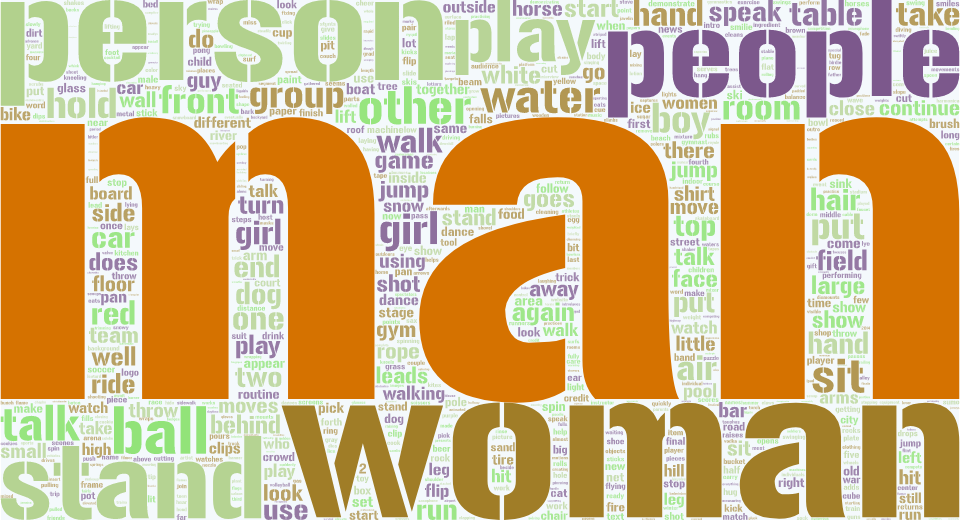}
    }\\
    \subfloat[{Distribution of videos} \lblfig{AS2}]{
        \hspace{0.1cm}
        \includegraphics[width=.4\linewidth]{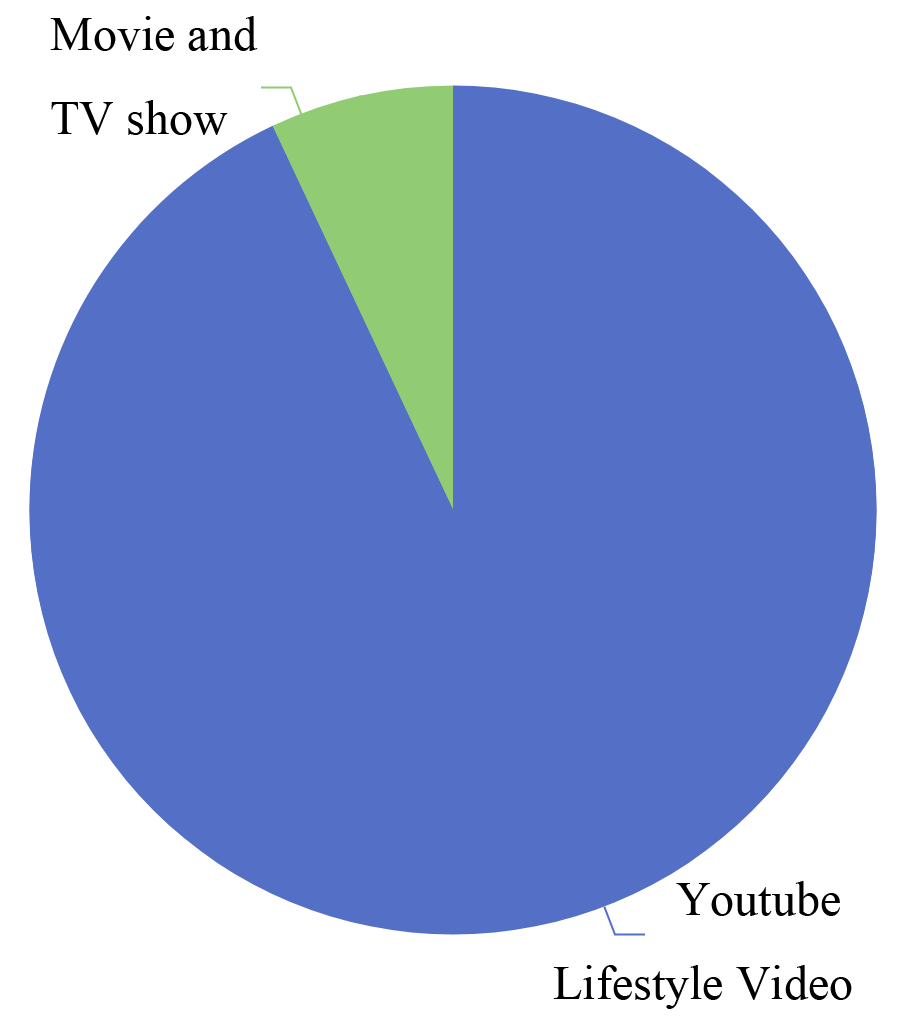}
    }
    \hspace{1cm}
    \subfloat[{Distribution of examples} \lblfig{AS3}]{
        \includegraphics[width=.4\linewidth]{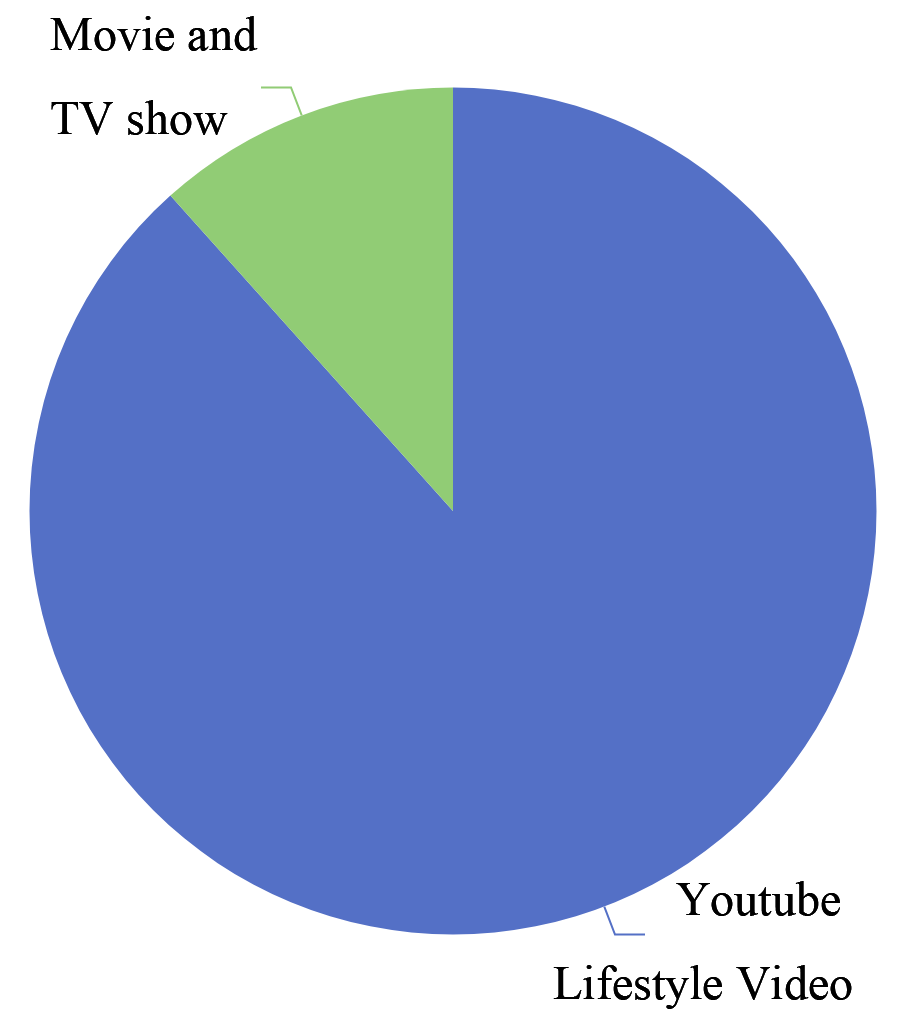}
    }\\
    \vspace{-4pt}
    \subfloat[{Distribution of premise events} \lblfig{AS4}]{
        \includegraphics[width=1.\linewidth]{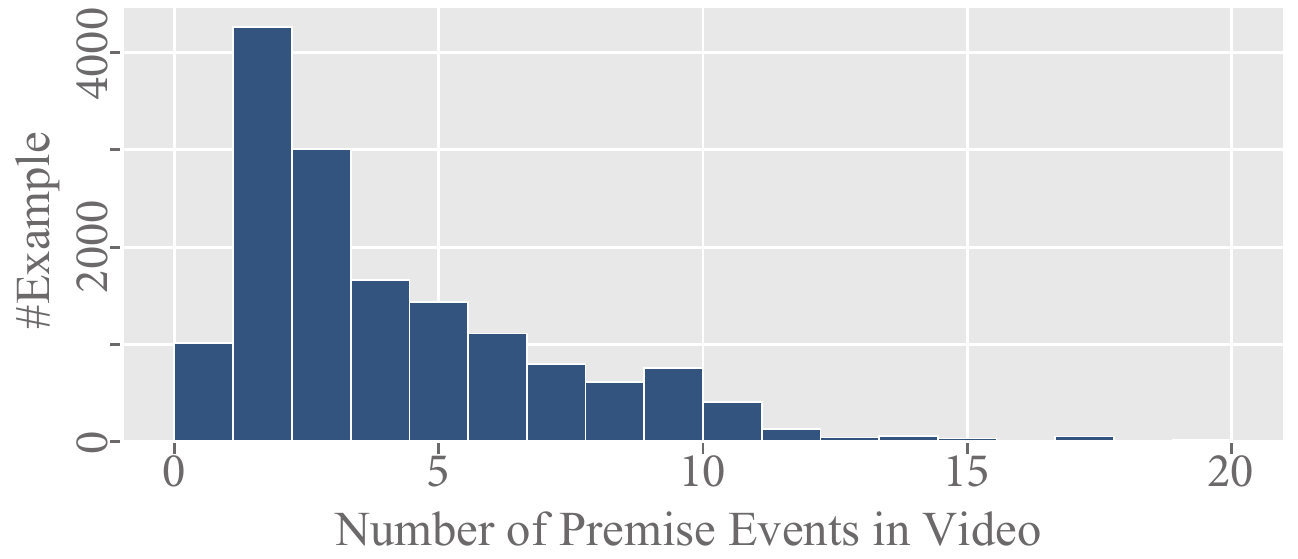}
    }
    \caption{\small Additional summative statistics of {\DATASET}  dataset (\S\ref{sec:statistic}).}
    \lblfig{statistic}
\end{figure}

\subsection{Detailed Annotation Process} \label{sec:ann}
In the annotation process, each video is passed at least four times:
(1) A first quick pass for filtering out videos without cause-effect relations;
(2) The second pass for annotating the event type, \ie, the premise and explanation.
In this phase, the entire video with initialized events and descriptions are all shown to the experts.
We use the original event annotation provided in~\cite{krishna2017dense,lei2020vlep,lei2020tvr} to initialize event boundaries, while they might contain noise annotations.
We thus request human experts to i) edit event boundaries when the initial separation can not well fit the description; ii) delete duplicate events when they are overlapped; iii) add additional events when they find a missing part in the cause-effect chain. And the annotation interface is shown in \figref{interface1_1}.
After that, experts are requested to further annotate the event type based on the events he selected in the previous step, as shown in \figref{interface1_2}.
(3) The third pass for abductive reasoning oriented description annotation.
Human experts are only shown with premise visual events while the hypothesis event is hidden. Annotation interfaces for annotating the premise and hypothesis are shown in \figref{interface2_1} and \figref{interface2_2} respectively.
Experts might vote to delete an example if they find that a plausible explanation can not be inferred from the premise. And the remaining examples are re-annotated with abductive reasoning oriented descriptions.
Notably, a video might contain multiple examples (candidate cause-effect chains), and we manually control the example distribution to make sure the same video will not be shown to the same expert twice.
(4) Forth, the final pass for validation. Both the annotated descriptions for premises and explanations for hypotheses are shown to another group of human experts. And they will vote for the validity.

\begin{figure}[t]
    \begin{center}
        \includegraphics[width=1.\linewidth]{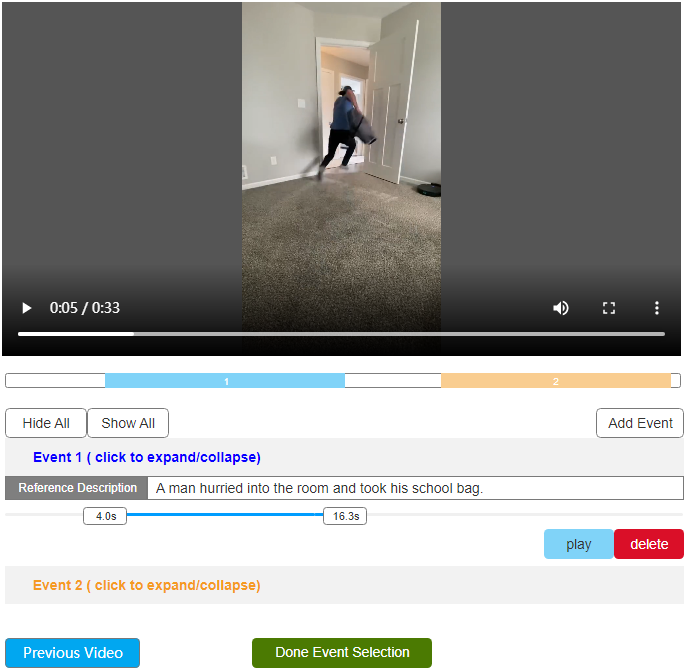}
    \end{center}
    \vspace{-3.3ex}
    \captionsetup{font=small}
    \caption{\small Interface for event selection. See \S\ref{sec:ann} for more details. }
    \label{fig:interface1_1}
    \vspace{-1ex}
\end{figure}

\begin{figure}[t]
    \begin{center}
        \includegraphics[width=1.\linewidth]{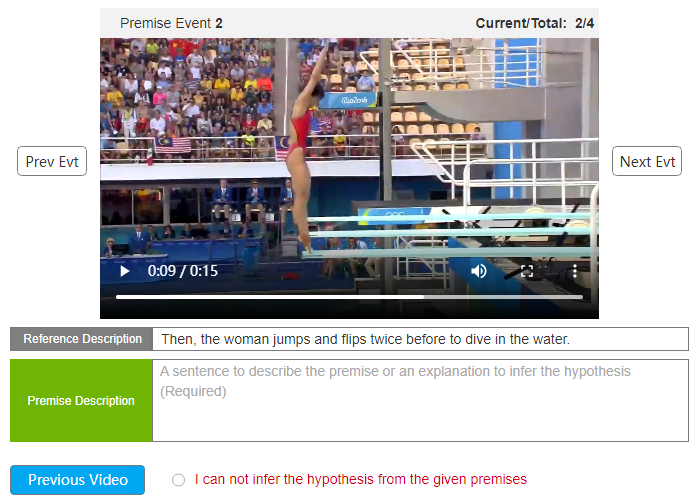}
    \end{center}
    \vspace{-2.5ex}
    \captionsetup{font=small}
    \caption{\small Interface for describing premises. Details are in \S\ref{sec:ann}. }
    \label{fig:interface2_1}
\end{figure}

\begin{figure}[t]
    \begin{center}
        \includegraphics[width=1.\linewidth]{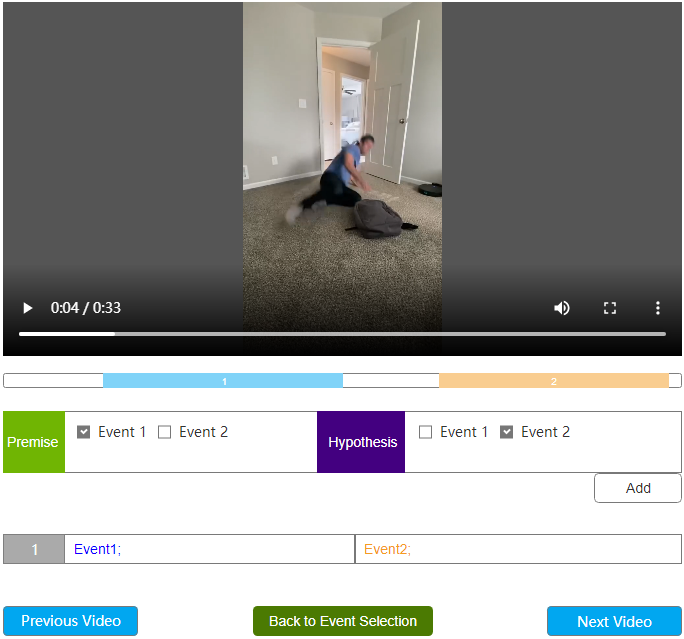}
    \end{center}
    \vspace{-1ex}
    \captionsetup{font=small}
    \caption{\small Interface for event type annotation. Details are in \S\ref{sec:ann}. }
    \label{fig:interface1_2}
    \vspace{-1ex}
\end{figure}

\begin{figure}[t]
    \begin{center}
        \includegraphics[width=1.\linewidth]{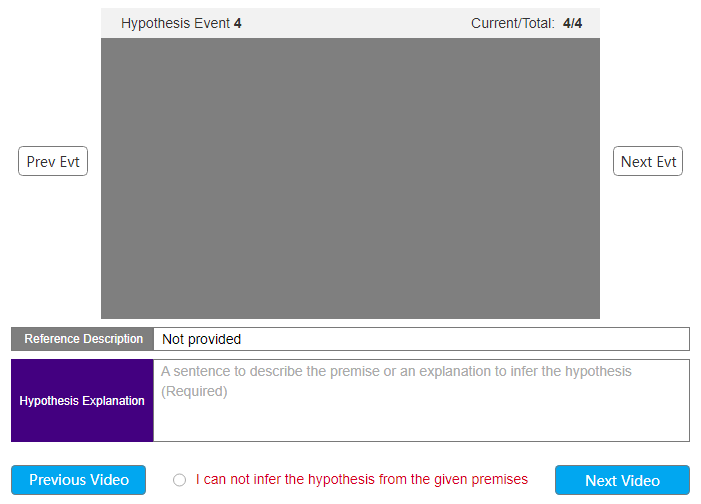}
    \end{center}
    \vspace{-12pt}
    \captionsetup{font=small}
    \caption{\small Interface{$_{\!}$} for{$_{\!}$} explaining{$_{\!}$} hypotheses.{$_{\!}$} Details{$_{\!}$} are{$_{\!}$} in{$_{\!}$} \S\ref{sec:ann}. }
    \label{fig:interface2_2}
\end{figure}

\section{Additional Details of Baselines}\label{sec:baseline}

We benchmark five top-leading DVC models~\cite{zhou2018end,dai2019transformer,lei2020mart,xiong2018move,wang2021end} on {\DATASET}. In this section, we detail the implementation and training protocol of these baseline methods.
\paragraph{MFT}
MFT~\cite{xiong2018move} is an LSTM-based method that consists of a selection LSTM for relevant event filtering and a captioning LSTM for coherent sentence generation.
We adapt it to {\TASKAc} task by recurrently passing the given events into MFT and the selection LSTM is transformed into a visually coherent maintaining module.
The unidirectionally casual structure can be captured that enables the inference on potential effect along the temporal order.

\paragraph{PDVC}
Similarly, PDVC~\cite{wang2021end} employs an LSTM-based captioning decoder, while it is conditioned on a deformable soft attention aggregated visual event.
And the visual events are embedded with a Transformer-based encoder that could also capture bidirectional causal dependencies within it.

\begin{table*}[!t]
    \centering\small
    \hspace{-.27cm}
    \resizebox{1.\textwidth}{!}{
        \setlength\tabcolsep{2pt}
        \renewcommand\arraystretch{1.1}
        \begin{tabular}{c|c||ccccc|ccccc}
            \hline\thickhline
            \rowcolor{mygray}
                                                       &                      & \multicolumn{5}{c|}{\textbf{Premise Event}} & \multicolumn{5}{c}{\textbf{Explanation Event}} \\
            \rowcolor{mygray}
            \multirow{-2}{*}{Method}&\multirow{-2}{*}{Setting} & BLEU@4 & METEOR & ROUGE & CIDEr & BERT-S & BLEU@4 & METEOR & ROUGE & CIDEr & BERT-S \\
            \hline \hline
            \multirow{4}{*}{\textsc{Reasoner}} &no hidden event  &
            5.26 & 11.32 & 24.94 & 39.52 & 35.09 &-&-&-&-&-  \\
            &with premise text only &-&-&-&-&-& 1.72 & 8.37 & 18.10 & 15.80 & 25.28 \\\cline{2-12}
            &\tabincell{c}{\textit{w/o} external knowledge\\(\textbf{reported in the main paper})}          & \textbf{5.03}& \textbf{10.75} & \textbf{24.81} & \textbf{38.27} & \textbf{34.88} & \textbf{3.44}& \textbf{9.05}& \textbf{22.89} & \textbf{30.75} & \textbf{30.64} \\
            \hline
        \end{tabular}
    }
    \captionsetup{font=small}
    \caption{\small \textbf{Additional quantitative results} on the  \texttt{test} set of our {\DATASET} dataset. See \S\ref{sec:sexp} for details.}
    \label{tab:exp}
	\vspace{-6pt}
\end{table*}

\paragraph{VTrans}
VTrans~\cite{zhou2018end} is a fully attentional model that originates from the vanilla Transformer proposed in~\cite{vaswani2017attention}.
We follow the implementation in~\cite{lei2020mart}, which serves as a baseline that only considers a single event and independently generates a single sentence describing the given event.
Thus causal structure can not be formulated in this method.

\paragraph{Trans-XL}
Transformer-XL (Trans-XL)~\cite{dai2019transformer} is originally proposed for modeling unlimited longer-term dependencies with a segment-level recurrent strategy.
It can capture the intrinsic unidirectional causal structure within recurrent steps.
Following the implementation in~\cite{lei2020mart}, gradients can flow through recurrent steps instead of being stopped. This enables stronger long-term modeling.

\paragraph{MART}
MART~\cite{lei2020mart} is also built on a fully Transformer-based encoder-decoder architecture, that maintains a summarized memory module to model dependencies among events.
Similar to Trans-XL, the unidirectional causal structure is preserved in the memory.
Whereas, experimental results show that it suffers more on our {\TASKAc} potentially due to the content drift brought by masked visual hypotheses.

MFT and PDVC are benchmarked following the original training protocols.
We adapt VTrans, Trans-XL and MART to the {\TASKAc} task with the implementation provided by~\cite{lei2020mart}.
All of these baselines are trained with given events from our {\DATASET} \texttt{train} and evaluated on {\DATASET} \texttt{test} under the same setting of {\OURS} as reported in our main paper.

\vspace{-2pt}
\section{Additional Experimental Results}\label{sec:sexp}
\textcolor{black}{
    To shed light on the essence of both the Visual Abductive Reasoning task and our VAR dataset, we study two edge cases of the main setting:
    \textbf{i)}  First, all events are made available to the model, so that no abductive reasoning is needed. And the VAR task is degraded to a basic Dense Video Captioning task.
    \textbf{ii)} Second, only ground-truth linguistic descriptions of premise events are supplied to the models. Therefore, models are expected to conduct abductive reasoning with and only with linguistic cues.
}

\textcolor{black}{
    In Table~\ref{tab:exp}, we summarize the quantitative results of these two settings. As seen, when there is no hidden event, \ie, the incomplete causal structure is directly provided, {\OURS} achieves even better performance. It reveals that fulfilling explanation events through abductive reasoning is indeed challenging and the causal structure understanding is also helpful to basic visual recognition.
    And for the next setting, when only premise texts are given, comparing to fully utilize both visual and linguistic cues, {\OURS} can not well-infer the hypothesis within linguistic modality only, which proves that the visual-based abductive reasoning is indispensable in the VAR task.
}

In our main paper, for the benchmarking results, we report the average scores of ten trained models with different random seeds. To prove the statistical significance of our results, here we further provide the corresponding standard deviations of {\OURS} and two representative methods~\cite{lei2020mart,wang2021end} in Table~\ref{tab:sta}.

\begin{table}[t]
	\centering
	\small
	\resizebox{1.\columnwidth}{!}{
        \hspace{-1ex}
		\setlength\tabcolsep{2pt}
		\renewcommand\arraystretch{1.1}
		\begin{tabular}{c||ccc|ccc}
			\hline\thickhline
            \rowcolor{mygray}
                                     & \multicolumn{3}{c|}{\textbf{Premise Event}} & \multicolumn{3}{c}{\textbf{Explanation Event}} \\
            \rowcolor{mygray}
            \multirow{-2}{*}{Method} & BLEU@4 & CIDEr & BERT-S & BLEU@4 & CIDEr & BERT-S \\\hline\hline
			{\cite{lei2020mart}}    & 3.74$\pm$0.07 & 29.22$\pm$0.39 & 29.53$\pm$0.12 & 2.86$\pm$0.07 & 24.05$\pm$0.27 & 27.77$\pm$0.16 \\
			{\cite{wang2021end}}    & 4.28$\pm$0.04 & 33.59$\pm$0.30 & 29.37$\pm$0.18 & 3.00$\pm$0.05 & 25.14$\pm$0.21 & 27.80$\pm$0.17 \\
            \textsc{Reasoner}       & 5.03$\pm$0.02 & 38.27$\pm$0.15 & 34.88$\pm$0.10 & 3.44$\pm$0.01 & 30.75$\pm$0.24 & 30.64$\pm$0.08 \\
			\hline
		\end{tabular}
	}
    \captionsetup{font=small}
    \caption{\small Average scores and their standard deviations of \textsc{Reasoner} and two representative methods \cite{lei2020mart,wang2021end} (\S\ref{sec:sexp}).}
    \label{tab:sta}
	\vspace{-2pt}
\end{table}

\section{Additional Qualitative Visualization}\label{sec:qualitative}
In \figref{qualitative}{failure}, we show more qualitative examples from {\DATASET} \texttt{test} following Fig.~7 in the main paper.
Generated sentences from competitors~\cite{wang2021end,lei2020mart,zhou2018end} along with our {\OURS} are presented in~\figref{qualitative}.
In contrast to the competitors, both adequate descriptions on premises and plausible inferences for hypotheses are observed for our proposed method, {\OURS}, which demonstrates a superior abductive reasoning ability for capturing causal structures among visual events.
Some failure cases and gold human-written explanations are shown in~\figref{failure}.
Even though {\OURS} shows impressive performance on inferring with abduction, {\TASKAc} task is still a mostly unsolved technical problem. And there remains a large headroom for future works to conquer.

\section{Limitation and Reproducibility}\label{sec:limitation}

\subsection{Dataset Limitation}\label{sec:datalim}
During annotation process, we observe a bias against women and minorities due to the highly biased nature of movie and web sourced video data~\cite{tang2021mitigating,sap2017connotation,rudinger2017social}.
{\DATASET}, derived from these data, inevitably runs into the same problem~\cite{zellers2019recognition,hendricks2018women}.
We thus suggest that models trained on {\DATASET} dataset should be cautiously examined before being deployed onto real-world applications.
And we will devote further efforts to mitigating the issue in our later works.

\subsection{Details of BERTScore Evaluation}
BERTScore~\cite{zhang2019bertscore} leverages the pre-trained contextual embeddings from BERT-based models for similarity measurement.
Thus the evaluated scores vary a lot with different model settings.
In this paper, all reported BERTScores are evaluated under a hash code version: roberta-large{\textunderscore}L17{\textunderscore}no-idf{\textunderscore}version{$=$}0.3.0(hug{\textunderscore}trans{$=$}2.3.0)-rescaled.
\textbf{We encourage later works to follow the same setting for a fair comparison.} A static version of BERTScore is released at: \url{https://github.com/leonnnop/VAR}.

\section{Legal and Ethical Considerations}\label{sec:legal}
\subsection{Asset License}
Videos in {\DATASET} dataset are collected from four main assets:
 \textbf{(1)} ActivityNet Captions~\cite{krishna2017dense}\footnote{\scriptsize \url{https://cs.stanford.edu/people/ranjaykrishna/densevid/}}, 2017 version, under CC-BY 4.0 license\footnote{\label{ccby4}\scriptsize \url{https://creativecommons.org/licenses/by/4.0/}};
 \textbf{(2)} VLEP~\cite{lei2020vlep}\footnote{\scriptsize \url{https://github.com/jayleicn/VideoLanguageFuturePred}}, 2020~version, under CC-BY 4.0 license\footnoteref{ccby4};
 \textbf{(3)} TVC~\cite{lei2020tvr}\footnote{\scriptsize \url{https://github.com/jayleicn/TVCaption}}, 2020 version, under CC-BY 4.0 license\footnoteref{ccby4};
 \textbf{(4)} MovieClips\footnote{\scriptsize \url{https://www.movieclips.com/}}, copyright {\copyright} 2021 Fandango. The site and services are available for non-commercial use. Detailed terms of use are available online\footnote{\scriptsize \url{https://www.fandango.com/policies/terms-of-use}}.
{\DATASET} dataset will be released under CC-BY 4.0 license\footnoteref{ccby4}, respecting the licences of all its videos.

\subsection{Concerns on Personal Data Collection}
{\DATASET} is annotated by human experts and we conduct user studies to evaluate the human-subjective generation quality.
All human experts are noticed that the annotation and evaluation will be used for academic research and individual consents are reached with signed agreements.
The annotation will not leak any personal information about the experts.

\subsection{Potential Societal Impact}
Endowing an AI system with human intelligence has long been dreamed by AI researchers, which could fundamentally change the experience of human-machine interaction.
{\TASKAc} takes an important step towards more human-like AI systems that are endowed with abductive reasoning ability,
while it might provoke concerns about disinformation, \eg, fabricating deceptive facts.
We encourage more technical researching efforts devoted to fake content detection, and at the same time,
we will organize a gated release of our dataset and model to prevent potentially malicious abuses.

\clearpage

\begin{figure*}[t]
    \centering
    \subfloat{
        \includegraphics[width=1.\linewidth]{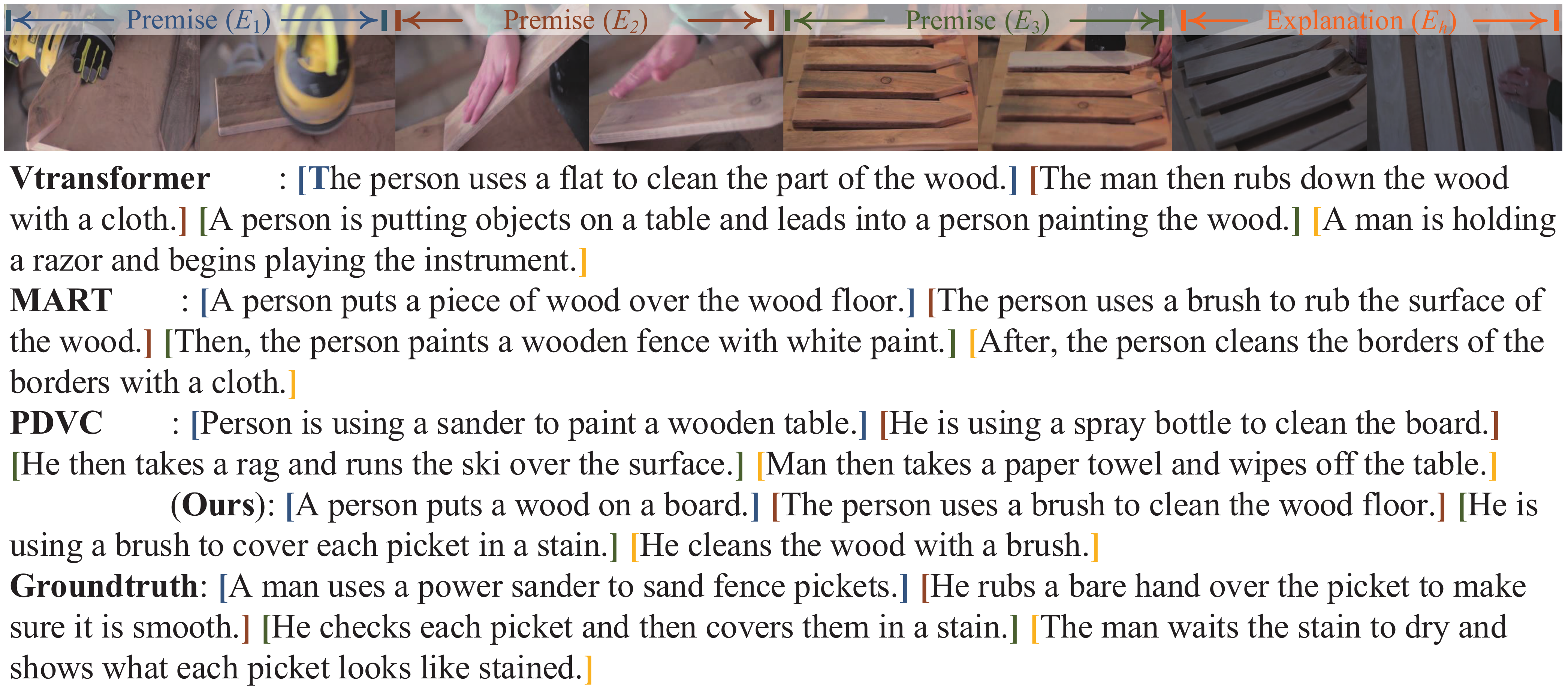}
        \put(-493.5, 55.5){\normalsize \textbf{\OURS}}
        \put(-462, 82){\normalsize \cite{wang2021end}}
        \put(-457, 120.5){\normalsize \cite{lei2020mart}}
        \put(-429, 159){\normalsize \cite{zhou2018end}}
    }\vspace{-2ex}
    \subfloat{
        \includegraphics[width=1.\linewidth]{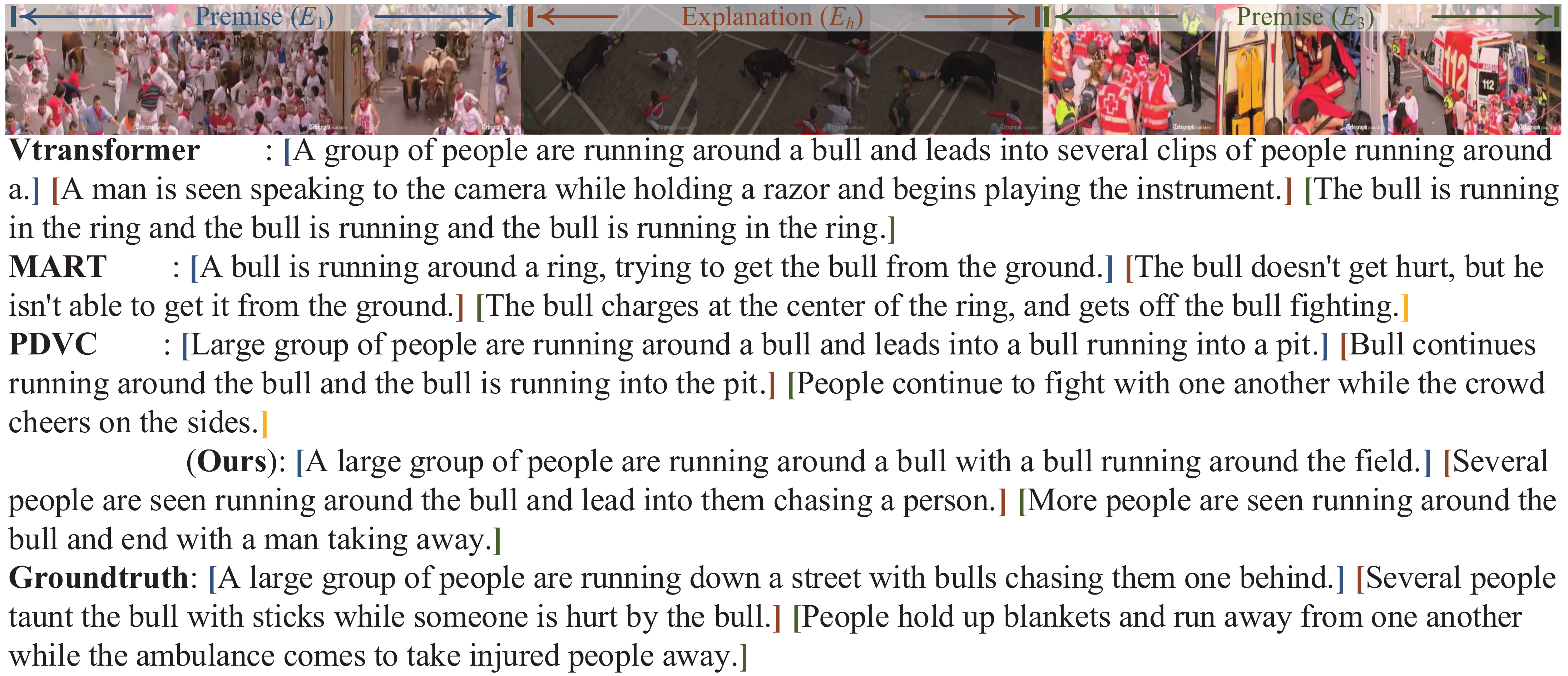}
        \put(-493.5, 65){\normalsize \textbf{\OURS}}
        \put(-464.5, 102.5){\normalsize \cite{wang2021end}}
        \put(-459, 127){\normalsize \cite{lei2020mart}}
        \put(-432.5, 163){\normalsize \cite{zhou2018end}}
    }
    \captionsetup{font=small}
    \caption{\small Additional qualitative comparisons of {\OURS} and \cite{wang2021end,lei2020mart,zhou2018end} on {\DATASET} \texttt{test}. See \S\ref{sec:qualitative} for more details. }
    \label{fig:qualitative}
    \vspace{-8pt}
\end{figure*}

\begin{figure*}[t]
    \begin{center}
        \includegraphics[width=1.\linewidth]{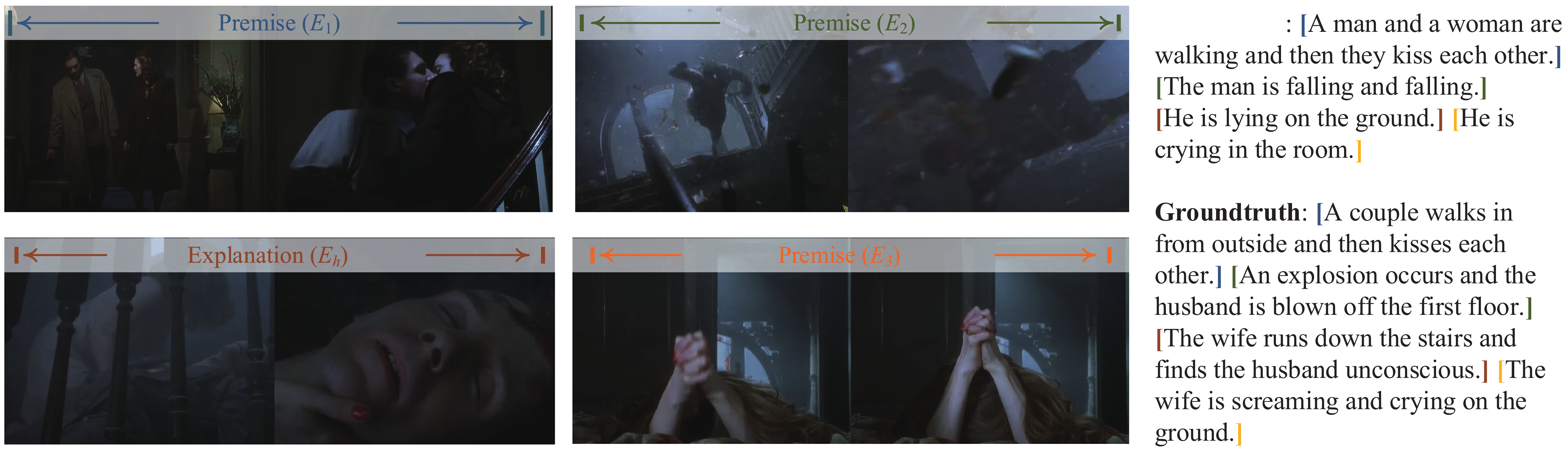}
        \put(-132, 132.5){\small \textbf{\OURS}}
    \end{center}
    \vspace{-12pt}
    \captionsetup{font=small}
    \caption{\small Additional failure cases of {\OURS} on {\DATASET} \texttt{test}. See \S\ref{sec:qualitative} for more details. }
    \label{fig:failure}
    \vspace{-4pt}
\end{figure*}
}

\clearpage
{\small
\bibliographystyle{ieee_fullname}
\bibliography{cite}

\begin{thebibliography}{10}\itemsep=-1pt

\bibitem{abu2018will}
Yazan Abu~Farha, Alexander Richard, and Juergen Gall.
\newblock When will you do what?-anticipating temporal occurrences of
  activities.
\newblock In {\em CVPR}, 2018.

\bibitem{alahi2016social}
Alexandre Alahi, Kratarth Goel, Vignesh Ramanathan, Alexandre Robicquet, Li
  Fei-Fei, and Silvio Savarese.
\newblock Social lstm: Human trajectory prediction in crowded spaces.
\newblock In {\em CVPR}, 2016.

\bibitem{banerjee2005meteor}
Satanjeev Banerjee and Alon Lavie.
\newblock Meteor: An automatic metric for mt evaluation with improved
  correlation with human judgments.
\newblock In {\em ACL}, 2005.

\bibitem{bengio2015scheduled}
Samy Bengio, Oriol Vinyals, Navdeep Jaitly, and Noam Shazeer.
\newblock Scheduled sampling for sequence prediction with recurrent neural
  networks.
\newblock In {\em NeurIPS}, 2015.

\bibitem{bhagavatula2020abductive}
Chandra Bhagavatula, Ronan~Le Bras, Chaitanya Malaviya, Keisuke Sakaguchi, Ari
  Holtzman, Hannah Rashkin, Doug Downey, Scott Wen-tau Yih, and Yejin Choi.
\newblock Abductive commonsense reasoning.
\newblock In {\em ICLR}, 2020.

\bibitem{caba2015activitynet}
Fabian Caba~Heilbron, Victor Escorcia, Bernard Ghanem, and Juan Carlos~Niebles.
\newblock Activitynet: A large-scale video benchmark for human activity
  understanding.
\newblock In {\em CVPR}, 2015.

\bibitem{chang2020procedure}
Chien-Yi Chang, De-An Huang, Danfei Xu, Ehsan Adeli, Li Fei-Fei, and
  Juan~Carlos Niebles.
\newblock Procedure planning in instructional videos.
\newblock In {\em ECCV}, 2020.

\bibitem{chu2021twins}
Xiangxiang Chu, Zhi Tian, Yuqing Wang, Bo Zhang, Haibing Ren, Xiaolin Wei,
  Huaxia Xia, and Chunhua Shen.
\newblock Twins: Revisiting spatial attention design in vision transformers.
\newblock In {\em NeurIPS}, 2021.

\bibitem{dai2019transformer}
Zihang Dai, Zhilin Yang, Yiming Yang, Jaime Carbonell, Quoc~V Le, and Ruslan
  Salakhutdinov.
\newblock Transformer-xl: Attentive language models beyond a fixed-length
  context.
\newblock In {\em ACL}, 2019.

\bibitem{deng2021sketch}
Chaorui Deng, Shizhe Chen, Da Chen, Yuan He, and Qi Wu.
\newblock Sketch, ground, and refine: Top-down dense video captioning.
\newblock In {\em CVPR}, 2021.

\bibitem{devlin2018bert}
Jacob Devlin, Ming-Wei Chang, Kenton Lee, and Kristina Toutanova.
\newblock Bert: Pre-training of deep bidirectional transformers for language
  understanding.
\newblock In {\em NAACL-HLT}, 2018.

\bibitem{dosovitskiy2020image}
Alexey Dosovitskiy, Lucas Beyer, Alexander Kolesnikov, Dirk Weissenborn,
  Xiaohua Zhai, Thomas Unterthiner, Mostafa Dehghani, Matthias Minderer, Georg
  Heigold, Sylvain Gelly, et~al.
\newblock An image is worth 16x16 words: Transformers for image recognition at
  scale.
\newblock In {\em ICLR}, 2021.

\bibitem{epstein2021learning}
Dave Epstein and Carl Vondrick.
\newblock Learning goals from failure.
\newblock In {\em CVPR}, 2021.

\bibitem{fragkiadaki2015recurrent}
Katerina Fragkiadaki, Sergey Levine, Panna Felsen, and Jitendra Malik.
\newblock Recurrent network models for human dynamics.
\newblock In {\em ICCV}, 2015.

\bibitem{he2016deep}
Kaiming He, Xiangyu Zhang, Shaoqing Ren, and Jian Sun.
\newblock Deep residual learning for image recognition.
\newblock In {\em CVPR}, 2016.

\bibitem{hendricks2018women}
Lisa~Anne Hendricks, Kaylee Burns, Kate Saenko, Trevor Darrell, and Anna
  Rohrbach.
\newblock Women also snowboard: Overcoming bias in captioning models.
\newblock In {\em ECCV}, 2018.

\bibitem{hesselhwang2022abduction}
Jack Hessel, Jena~D Hwang, Jae~Sung Park, Rowan Zellers, Chandra Bhagavatula,
  Anna Rohrbach, Kate Saenko, and Yejin Choi.
\newblock The abduction of sherlock holmes: A dataset for visual abductive
  reasoning.
\newblock {\em arXiv preprint arXiv:2202.04800}, 2022.

\bibitem{hong2021transformation}
Xin Hong, Yanyan Lan, Liang Pang, Jiafeng Guo, and Xueqi Cheng.
\newblock Transformation driven visual reasoning.
\newblock In {\em CVPR}, 2021.

\bibitem{huang2020inset}
Yichen Huang, Yizhe Zhang, Oussama Elachqar, and Yu Cheng.
\newblock Inset: Sentence infilling with inter-sentential transformer.
\newblock In {\em ACL}, 2020.

\bibitem{ioffe2015batch}
Sergey Ioffe and Christian Szegedy.
\newblock Batch normalization: Accelerating deep network training by reducing
  internal covariate shift.
\newblock In {\em ICML}, 2015.

\bibitem{ippolito2019unsupervised}
Daphne Ippolito, David Grangier, Chris Callison-Burch, and Douglas Eck.
\newblock Unsupervised hierarchical story infilling.
\newblock In {\em Proceedings of the First Workshop on Narrative
  Understanding}, 2019.

\bibitem{jain2016structural}
Ashesh Jain, Amir~R Zamir, Silvio Savarese, and Ashutosh Saxena.
\newblock Structural-rnn: Deep learning on spatio-temporal graphs.
\newblock In {\em CVPR}, 2016.

\bibitem{ji2021hierarchical}
Lei Ji, Xianglin Guo, Haoyang Huang, and Xilin Chen.
\newblock Hierarchical context-aware network for dense video event captioning.
\newblock In {\em ACL}, 2021.

\bibitem{kang2019linguistic}
Dongyeop Kang and Eduard Hovy.
\newblock Linguistic versus latent relations for modeling coherent flow in
  paragraphs.
\newblock In {\em EMNLP-IJCNLP}, 2019.

\bibitem{ke2020rethinking}
Guolin Ke, Di He, and Tie-Yan Liu.
\newblock Rethinking positional encoding in language pre-training.
\newblock In {\em ICLR}, 2020.

\bibitem{keil2003folkscience}
Frank~C Keil.
\newblock Folkscience: Coarse interpretations of a complex reality.
\newblock {\em Trends in cognitive sciences}, 2003.

\bibitem{keil2006explanation}
Frank~C Keil.
\newblock Explanation and understanding.
\newblock {\em Annual review of psychology}, 2006.

\bibitem{kitani2012activity}
Kris~M Kitani, Brian~D Ziebart, James~Andrew Bagnell, and Martial Hebert.
\newblock Activity forecasting.
\newblock In {\em ECCV}, 2012.

\bibitem{kojima2002natural}
Atsuhiro Kojima, Takeshi Tamura, and Kunio Fukunaga.
\newblock Natural language description of human activities from video images
  based on concept hierarchy of actions.
\newblock {\em IJCV}, 2002.

\bibitem{krishna2017dense}
Ranjay Krishna, Kenji Hata, Frederic Ren, Li Fei-Fei, and Juan Carlos~Niebles.
\newblock Dense-captioning events in videos.
\newblock In {\em ICCV}, 2017.

\bibitem{lan2014hierarchical}
Tian Lan, Tsung-Chuan Chen, and Silvio Savarese.
\newblock A hierarchical representation for future action prediction.
\newblock In {\em ECCV}, 2014.

\bibitem{lei2020mart}
Jie Lei, Liwei Wang, Yelong Shen, Dong Yu, Tamara~L Berg, and Mohit Bansal.
\newblock Mart: Memory-augmented recurrent transformer for coherent video
  paragraph captioning.
\newblock In {\em ACL}, 2020.

\bibitem{lei2020tvr}
Jie Lei, Licheng Yu, Tamara~L Berg, and Mohit Bansal.
\newblock Tvr: A large-scale dataset for video-subtitle moment retrieval.
\newblock In {\em ECCV}, 2020.

\bibitem{lei2020vlep}
Jie Lei, Licheng Yu, Tamara~L Berg, and Mohit Bansal.
\newblock What is more likely to happen next? video-and-language future event
  prediction.
\newblock In {\em EMNLP}, 2020.

\bibitem{li2018jointly}
Yehao Li, Ting Yao, Yingwei Pan, Hongyang Chao, and Tao Mei.
\newblock Jointly localizing and describing events for dense video captioning.
\newblock In {\em CVPR}, 2018.

\bibitem{lin2002manual}
Chin-Yew Lin and Eduard Hovy.
\newblock Manual and automatic evaluation of summaries.
\newblock In {\em ACL}, 2002.

\bibitem{liu2021video}
Hui Liu and Xiaojun Wan.
\newblock Video paragraph captioning as a text summarization task.
\newblock In {\em ACL}, 2021.

\bibitem{logan2018topology}
RK Logan and PO Izabella.
\newblock A topology of mind -- spiral thought patterns, the hyperlinking of
  text, ideas and more, 2018.

\bibitem{logan2018thinking}
Robert~K Logan and Marlie Tandoc.
\newblock Thinking in patterns and the pattern of human thought as contrasted
  with ai data processing.
\newblock {\em Information}, 9(4):83, 2018.

\bibitem{lombrozoexplanation}
Tania Lombrozo.
\newblock Explanation and abductive inference.
\newblock {\em The Oxford Handbook of Thinking and Reasoning}.

\bibitem{martinez2017human}
Julieta Martinez, Michael~J Black, and Javier Romero.
\newblock On human motion prediction using recurrent neural networks.
\newblock In {\em CVPR}, 2017.

\bibitem{mathieu2016deep}
Michael Mathieu, Camille Couprie, and Yann LeCun.
\newblock Deep multi-scale video prediction beyond mean square error.
\newblock In {\em ICLR}, 2016.

\bibitem{melas2018training}
Luke Melas-Kyriazi, Alexander~M Rush, and George Han.
\newblock Training for diversity in image paragraph captioning.
\newblock In {\em EMNLP}, 2018.

\bibitem{mun2019streamlined}
Jonghwan Mun, Linjie Yang, Zhou Ren, Ning Xu, and Bohyung Han.
\newblock Streamlined dense video captioning.
\newblock In {\em CVPR}, 2019.

\bibitem{pan2020spatial}
Boxiao Pan, Haoye Cai, De-An Huang, Kuan-Hui Lee, Adrien Gaidon, Ehsan Adeli,
  and Juan~Carlos Niebles.
\newblock Spatio-temporal graph for video captioning with knowledge
  distillation.
\newblock In {\em CVPR}, 2020.

\bibitem{pan2016hierarchical}
Pingbo Pan, Zhongwen Xu, Yi Yang, Fei Wu, and Yueting Zhuang.
\newblock Hierarchical recurrent neural encoder for video representation with
  application to captioning.
\newblock In {\em CVPR}, 2016.

\bibitem{papineni2002bleu}
Kishore Papineni, Salim Roukos, Todd Ward, and Wei-Jing Zhu.
\newblock Bleu: a method for automatic evaluation of machine translation.
\newblock In {\em ACL}, 2002.

\bibitem{park2019robust}
Dong~Huk Park, Trevor Darrell, and Anna Rohrbach.
\newblock Robust change captioning.
\newblock In {\em ICCV}, 2019.

\bibitem{park2020visualcomet}
Jae~Sung Park, Chandra Bhagavatula, Roozbeh Mottaghi, Ali Farhadi, and Yejin
  Choi.
\newblock Visualcomet: Reasoning about the dynamic context of a still image.
\newblock In {\em ECCV}, 2020.

\bibitem{park2019adversarial}
Jae~Sung Park, Marcus Rohrbach, Trevor Darrell, and Anna Rohrbach.
\newblock Adversarial inference for multi-sentence video description.
\newblock In {\em CVPR}, 2019.

\bibitem{peirce1931collected}
Charles~Sanders Peirce.
\newblock {\em Collected papers of charles sanders peirce}.
\newblock Harvard University Press, 1931.

\bibitem{qin2019counterfactual}
Lianhui Qin, Antoine Bosselut, Ari Holtzman, Chandra Bhagavatula, Elizabeth
  Clark, and Yejin Choi.
\newblock Counterfactual story reasoning and generation.
\newblock In {\em EMNLP-IJCNLP}, 2019.

\bibitem{raffel2020exploring}
Colin Raffel, Noam Shazeer, Adam Roberts, Katherine Lee, Sharan Narang, Michael
  Matena, Yanqi Zhou, Wei Li, and Peter~J Liu.
\newblock Exploring the limits of transfer learning with a unified text-to-text
  transformer.
\newblock {\em JMLR}, 2020.

\bibitem{rudinger2017social}
Rachel Rudinger, Chandler May, and Benjamin Van~Durme.
\newblock Social bias in elicited natural language inferences.
\newblock In {\em ACL workshops}, 2017.

\bibitem{ryoo2011human}
Michael~S Ryoo.
\newblock Human activity prediction: Early recognition of ongoing activities
  from streaming videos.
\newblock In {\em ICCV}, 2011.

\bibitem{sap2017connotation}
Maarten Sap, Marcella~Cindy Prasettio, Ari Holtzman, Hannah Rashkin, and Yejin
  Choi.
\newblock Connotation frames of power and agency in modern films.
\newblock In {\em EMNLP}, 2017.

\bibitem{shanahan2005perception}
Murray Shanahan.
\newblock Perception as abduction: Turning sensor data into meaningful
  representation.
\newblock {\em Cognitive science}, 2005.

\bibitem{shaw2018self}
Peter Shaw, Jakob Uszkoreit, and Ashish Vaswani.
\newblock Self-attention with relative position representations.
\newblock In {\em NAACL-HLT}, 2018.

\bibitem{shelley1995visual}
Cameron Shelley.
\newblock Visual abduction in anthropology and archaeology.
\newblock In {\em Systematic Methods of Scientific Discovery: Papers from the
  1995 Spring Symposium}, 1995.

\bibitem{song2021towards}
Yuqing Song, Shizhe Chen, and Qin Jin.
\newblock Towards diverse paragraph captioning for untrimmed videos.
\newblock In {\em CVPR}, 2021.

\bibitem{sun2019relational}
Chen Sun, Abhinav Shrivastava, Carl Vondrick, Rahul Sukthankar, Kevin Murphy,
  and Cordelia Schmid.
\newblock Relational action forecasting.
\newblock In {\em CVPR}, 2019.

\bibitem{suris2021learning}
D{\'\i}dac Sur{\'\i}s, Ruoshi Liu, and Carl Vondrick.
\newblock Learning the predictability of the future.
\newblock In {\em CVPR}, 2021.

\bibitem{tang2021mitigating}
Ruixiang Tang, Mengnan Du, Yuening Li, Zirui Liu, Na Zou, and Xia Hu.
\newblock Mitigating gender bias in captioning systems.
\newblock In {\em WWW}, 2021.

\bibitem{taylor1953cloze}
Wilson~L Taylor.
\newblock ``cloze procedure'': A new tool for measuring readability.
\newblock {\em Journalism quarterly}, 30(4):415--433, 1953.

\bibitem{vaswani2017attention}
Ashish Vaswani, Noam Shazeer, Niki Parmar, Jakob Uszkoreit, Llion Jones,
  Aidan~N Gomez, Lukasz Kaiser, and Illia Polosukhin.
\newblock Attention is all you need.
\newblock In {\em NeurIPS}, 2017.

\bibitem{vedantam2015cider}
Ramakrishna Vedantam, C Lawrence~Zitnick, and Devi Parikh.
\newblock Cider: Consensus-based image description evaluation.
\newblock In {\em CVPR}, 2015.

\bibitem{venugopalan2015sequence}
Subhashini Venugopalan, Marcus Rohrbach, Jeffrey Donahue, Raymond Mooney,
  Trevor Darrell, and Kate Saenko.
\newblock Sequence to sequence-video to text.
\newblock In {\em ICCV}, 2015.

\bibitem{venugopalan2015translating}
Subhashini Venugopalan, Huijuan Xu, Jeff Donahue, Marcus Rohrbach, Raymond~J
  Mooney, and Kate Saenko.
\newblock Translating videos to natural language using deep recurrent neural
  networks.
\newblock In {\em NAACL-HLT}, 2015.

\bibitem{vondrick2016anticipating}
Carl Vondrick, Hamed Pirsiavash, and Antonio Torralba.
\newblock Anticipating visual representations from unlabeled video.
\newblock In {\em CVPR}, 2016.

\bibitem{vondrick2016generating}
Carl Vondrick, Hamed Pirsiavash, and Antonio Torralba.
\newblock Generating videos with scene dynamics.
\newblock {\em NIPS}, 2016.

\bibitem{wang2021end}
Teng Wang, Ruimao Zhang, Zhichao Lu, Feng Zheng, Ran Cheng, and Ping Luo.
\newblock End-to-end dense video captioning with parallel decoding.
\newblock In {\em ICCV}, 2021.

\bibitem{williams1989learning}
Ronald~J Williams and David Zipser.
\newblock A learning algorithm for continually running fully recurrent neural
  networks.
\newblock {\em Neural computation}, 1989.

\bibitem{wu2021rethinking}
Kan Wu, Houwen Peng, Minghao Chen, Jianlong Fu, and Hongyang Chao.
\newblock Rethinking and improving relative position encoding for vision
  transformer.
\newblock In {\em ICCV}, 2021.

\bibitem{xiong2018move}
Yilei Xiong, Bo Dai, and Dahua Lin.
\newblock Move forward and tell: A progressive generator of video descriptions.
\newblock In {\em ECCV}, 2018.

\bibitem{yang2021multiple}
Yi Yang, Yueting Zhuang, and Yunhe Pan.
\newblock Multiple knowledge representation for big data artificial
  intelligence: framework, applications, and case studies.
\newblock {\em Frontiers of Information Technology \& Electronic Engineering},
  2021.

\bibitem{yao2015describing}
Li Yao, Atousa Torabi, Kyunghyun Cho, Nicolas Ballas, Christopher Pal, Hugo
  Larochelle, and Aaron Courville.
\newblock Describing videos by exploiting temporal structure.
\newblock In {\em ICCV}, 2015.

\bibitem{zellers2019recognition}
Rowan Zellers, Yonatan Bisk, Ali Farhadi, and Yejin Choi.
\newblock From recognition to cognition: Visual commonsense reasoning.
\newblock In {\em CVPR}, 2019.

\bibitem{zhang2018cross}
Bowen Zhang, Hexiang Hu, and Fei Sha.
\newblock Cross-modal and hierarchical modeling of video and text.
\newblock In {\em ECCV}, 2018.

\bibitem{zhang2019bertscore}
Tianyi Zhang, Varsha Kishore, Felix Wu, Kilian~Q Weinberger, and Yoav Artzi.
\newblock Bertscore: Evaluating text generation with bert.
\newblock In {\em ICLR}, 2019.

\bibitem{zhang2020object}
Ziqi Zhang, Yaya Shi, Chunfeng Yuan, Bing Li, Peijin Wang, Weiming Hu, and
  Zheng-Jun Zha.
\newblock Object relational graph with teacher-recommended learning for video
  captioning.
\newblock In {\em CVPR}, 2020.

\bibitem{zhou2019grounded}
Luowei Zhou, Yannis Kalantidis, Xinlei Chen, Jason~J Corso, and Marcus
  Rohrbach.
\newblock Grounded video description.
\newblock In {\em CVPR}, 2019.

\bibitem{zhou2018end}
Luowei Zhou, Yingbo Zhou, Jason~J Corso, Richard Socher, and Caiming Xiong.
\newblock End-to-end dense video captioning with masked transformer.
\newblock In {\em CVPR}, 2018.

\bibitem{zhu2019text}
Wanrong Zhu, Zhiting Hu, and Eric Xing.
\newblock Text infilling.
\newblock {\em arXiv preprint arXiv:1901.00158}, 2019.

\end{thebibliography}
}

\end{document}